\documentclass[10pt,twocolumn,letterpaper]{article}

\usepackage[pagenumbers]{cvpr} 

\usepackage{graphicx}
\usepackage{amsmath}
\usepackage{amssymb}
\usepackage{mathtools}
\usepackage{booktabs}
\usepackage{caption}
\usepackage{subcaption}
\usepackage{float}
\usepackage{enumitem}
\usepackage[percent]{overpic}
\usepackage[pagebackref,breaklinks,colorlinks]{hyperref}

\usepackage[capitalize]{cleveref}
\crefname{section}{Sec.}{Secs.}
\Crefname{section}{Section}{Sections}
\Crefname{table}{Table}{Tables}
\crefname{table}{Tab.}{Tabs.}

\def\cvprPaperID{1049} 
\def\confName{CVPR}
\def\confYear{2023}

\newif\ifdraft
\draftfalse
\drafttrue

\usepackage[table]{xcolor}
\definecolor{orange}{rgb}{1,0.5,0}
\definecolor{violet}{RGB}{70,0,170}

\ifdraft
 \newcommand{\PF}[1]{{\color{red}{\bf PF: #1}}}
 
 \newcommand{\BG}[1]{{\color{blue}{\bf BG: #1}}}
 
 \newcommand{\RL}[1]{{\color{orange}{\bf RL: #1}}}
 
 \newcommand{\LD}[1]{{\color{olive}{\bf LD: #1}}}
 
 \newcommand{\MS}[1]{{\color{pink}{\bf MS: #1}}}
 
\else
 \newcommand{\PF}[1]{}
 
 \newcommand{\BG}[1]{}
 
 \newcommand{\RL}[1]{}
 
 \newcommand{\LD}[1]{}
 
 \newcommand{\MS}[1]{}
 
\fi

\newcommand{\algoname}{{\tt DrapeNet}}
\newcommand{\bx}{\mathbf{x}}
\newcommand{\bz}{\mathbf{z}}

\newcommand\blfootnote[1]{%
  \begingroup
  \renewcommand\thefootnote{}\footnote{#1}%
  \addtocounter{footnote}{-1}%
  \endgroup
}

\begin{document}

\title{DrapeNet: Garment Generation and Self-Supervised Draping}

\author{
  Luca De Luigi$^{*,2}$ \\
  \and
  Ren Li$^{*,1}$\\
  \and
  Benoît Guillard$^{1}$\\
  \and
  Mathieu Salzmann$^{1}$\\
  \and 
  Pascal Fua$^{1}$
  \and
  \vspace{-1.1cm}\\
  {\small $^{1}$: CVLab, EPFL, \texttt{\{name.surname\}@epfl.ch}}
  {\small $^{2}$: University of Bologna, \texttt{luca.deluigi4@unibo.it}}
}


\twocolumn[{
\renewcommand\twocolumn[1][]{#1}
\maketitle
\begin{center}
    \vspace{-1cm}
    \begin{tabular}{c}
        \begin{overpic}[width=0.9\textwidth]{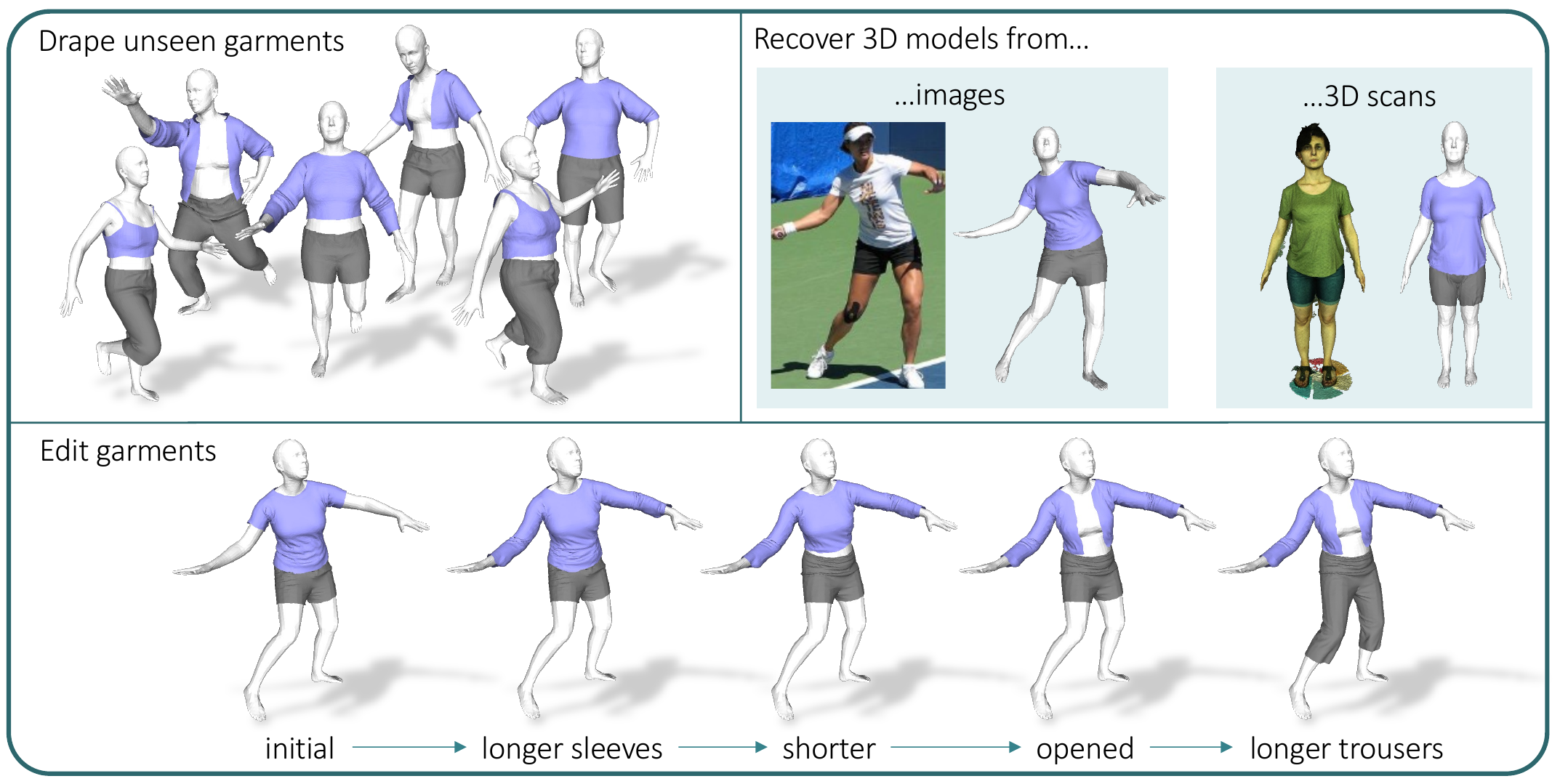}
        \end{overpic} \\
    \end{tabular}
\end{center}
\vspace{-0.45cm}
\small \hypertarget{fig:teaser}{Figure 1. Our network can drape garments over bodies of different shapes in various poses. To minimize the required amount of supervision, our draping network is trained with physics-based self-supervision and generalizes to multiple garments by being conditioned on latent codes. These can be manipulated to edit specific features of the corresponding garments. Being fully differentiable, our pipeline makes it possible to recover 3D models of garments and bodies from observations such as images and 3D scans.} 
\vspace{0.3cm}
}]

\blfootnote{$^{*}$ Equal contributions}
\vspace{-0.3cm}

\begin{abstract}

Recent approaches to drape garments quickly over arbitrary human bodies leverage self-supervision to eliminate the need for large training sets. However, they are designed to train one network per clothing item, which severely limits their generalization abilities. In our work, we rely on self-supervision to train a single network to drape multiple garments. This is achieved by predicting a 3D deformation field conditioned on the latent codes of a generative network, which models garments as unsigned distance fields.

Our pipeline can generate and drape previously unseen garments of any topology, whose shape can be edited by manipulating their latent codes. 
Being fully differentiable, our formulation makes it possible to recover accurate 3D models of garments from partial observations -- images or 3D scans -- via gradient descent. Our code is publicly available at \url{https://github.com/liren2515/DrapeNet}.

\end{abstract}


\section{Introduction}
\label{sec:intro}

Draping digital garments over differently-shaped bodies in random poses has been extensively studied due to its many applications such as fashion design, moviemaking, video gaming, virtual try-on and, nowadays, virtual and augmented reality. Physics-based simulation (PBS)~\cite{Baraff98,Liu17b,Provot95,Provot97,Tang13b,Vassilev01,zeller05,Nvcloth,Optitext,NvFlex,MarvelousDesigner,Su18b} can produce outstanding results, but at a high computational cost.

Recent years have witnessed the emergence of deep neural networks aiming to achieve the quality of PBS draping while being much faster, easily differentiable, and offering new speed vs. accuracy tradeoffs~\cite{Gundogdu19,Gundogdu22,Ma20,Patel20,Santesteban19,Shen20,Tiwari20,Viduarre20,Wang18f}. These networks are often trained to produce garments that resemble ground-truth ones. While effective, this requires building training datasets, consisting of ground-truth meshes obtained either from computationally expensive simulations~\cite{Narain12} or using  complex 3D scanning setups \cite{Pons-Moll17}. Moreover, to generalize to unseen garments and poses, these supervised approaches require training databases encompassing a great variety of samples depicting many combinations of garments, bodies and poses.

The recent PBNS and SNUG approaches~\cite{Santesteban22,Bertiche21b} address this by casting the physical models adopted in PBS into constraints used for self-supervision of deep learning models. This makes it possible to train the network on a multitude of body shapes and poses without ground-truth draped garments. Instead, the predicted garments are constrained to obey physics-based rules. However, both PBNS and SNUG, require training a separate network for each garment. They rely on mesh templates for garment representation and feature one output per mesh vertex. Thus, they cannot handle meshes with different topologies, even for the same garment. This makes them very specialized and limits their applicability to large garment collections as a new network must be trained for each new clothing item.

In this work, we introduce \algoname{}, an approach that also relies on physics-based constraints to provide self-supervision but can handle generic garments by conditioning a {\it single} draping network with a latent code describing the garment to be draped. We achieve this by coupling the draping network with a garment \textit{generative} network, composed of an encoder and a decoder. The encoder is trained to compress input garments into compact latent codes that are used as input condition for the draping network. The decoder, instead, reconstructs a 3D garment model from its latent code, thus allowing us to sample and edit new garments from the learned latent space.

Specifically, we model the output of the garment decoder as an unsigned distance function (UDF), which were demonstrated \cite{Guillard22b} to yield better accuracy and fewer interpenetrations than the inflated signed distance functions often used for this purpose \cite{Corona21,Li22c}. Moreover, UDFs can be triangulated in a differentiable way \cite{Guillard22b} to produce explicit surfaces that can easily be post-processed, making our pipeline fully differentiable.  Hence, \algoname{} can not only drape garments over given body shapes but can also perform gradient-based optimization to fit garments, along with body shapes and poses, to partial observations of clothed people, such as images or 3D scans.

Our contributions are as follows:
\begin{itemize}[noitemsep,topsep=0pt]
    \item We introduce a \textit{single} garment draping network conditioned on a  latent code to handle generic garments from a large collection (e.g. \textit{top} or \textit{bottom} garments);
    \item By exploiting physics-based self-supervision, our pipeline only requires a few hundred garment meshes in a canonical pose for training;
    \item Our framework enables the fast draping of new garments with high fidelity, as well as the sampling and editing of new garments from the learned latent space;
    \item Being fully differentiable, our method can be used to recover accurate 3D models of clothed people from images and 3D scans.
\end{itemize}


\section{Related Work}
\label{sec:related}

\textbf{Implicit Neural Representations for 3D Surfaces.}
Implicit neural representations have emerged a few years ago as an effective tool to represent surfaces whose topology is not known a priori. They can be implemented using (clipped) \textit{signed distance functions} (SDF)~\cite{Park19c} or \textit{occupancies}~\cite{Mescheder19}. When an explicit representation is required, it can be obtained using Marching Cubes~\cite{Lewiner03} and this can be done while preserving differentiability~ \cite{Remelli20b,Atzmon19,Mehta22}. However, they can only represent watertight surfaces. 

Thus, to represent open surfaces, such as clothes, it is possible to use inflated SDFs surrounding them.  However, this entails a loss in accuracy and there has been a recent push to replace SDFs by \textit{unsigned distance functions} (UDFs)~\cite{Chibane20b,Zhao21a,Venkatesh21}. One difficulty in so doing was that Marching Cubes was not designed with UDFs in mind, and obtaining explicit surfaces from these UDFs was therefore non-trivial. This has been addressed in~\cite{Guillard22b} by modifying the Marching Cubes algorithm to operate with UDFs. We model garment with UDFs and use ~\cite{Guillard22b} to mesh them. Other works augment signed distance fields with covariant fields to encode open surface garments~\cite{Santesteban22b,Buffet19}.


\begin{figure*}
    \centering
    \includegraphics[width=0.99\textwidth]{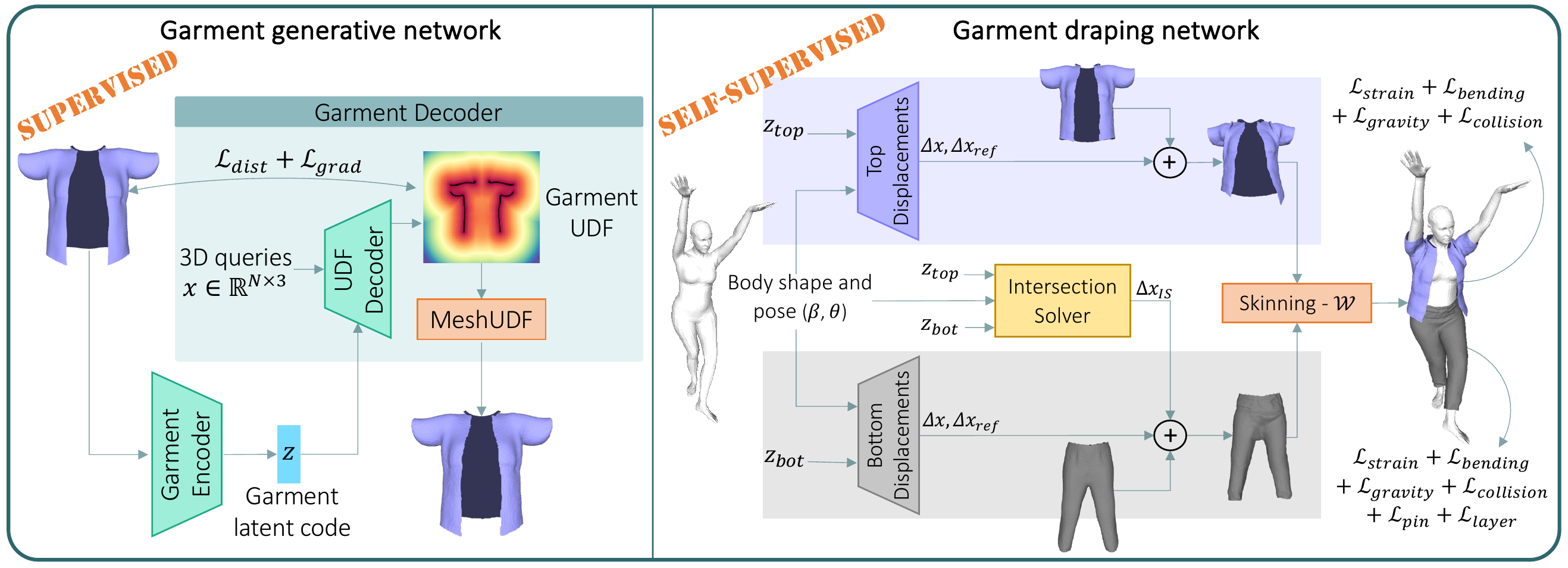}
    \vspace{-0.25cm}
    \caption{\textbf{Overview of our framework.} \textbf{Left:} Garment generative network, trained to embed garments into compact latent codes and predict their unsigned distance field (UDF) from such vectors. UDFs are then meshed using~\cite{Guillard22b}. \textbf{Right:} Garment draping network, conditioned on the latent codes of the generative network. It is trained in a self-supervised way to predict the displacements $\Delta x$ and $\Delta x_{\rm ref}$ to be applied to the vertices of given garments, before skinning them according to body shape and pose ($\beta$, $\theta$) with the predicted blending weights $\mathcal{W}$. It includes an Intersection Solver module to prevent intersection between top and bottom garments.}
    \label{fig:framework}
    \vspace{-0.35cm}
\end{figure*}

\textbf{Draping Garments over 3D Bodies.}
Two main classes of methods coexist, physics-based algorithms \cite{Baraff98,Li21h,Liang19a,Narain12,Narain13,Su18b} that produce high-quality drapings but at a high computational cost, and data-driven approaches that are faster but often at the cost of realism.

Among the latter, template-based approaches~\cite{Bertiche21b,Bhatnagar19,Jiang20d,Patel20,Santesteban21,Santesteban22,Tiwari20,Pan22} are dominant. Each garment is modeled by a specific triangulated mesh and a draping function is learned for each one. In other words, they do not generalize. There are however a number of exceptions. 
In~\cite{Gundogdu22,Bertiche21} the mesh is replaced by 3D point clouds that can represent generic garments. This enables deforming garments with arbitrary topology and geometric complexity, by estimating the deformation separately for each point. \cite{Zakharkin21} goes further and allows differentiable changes in garment topology by sampling a fixed number of points from the body mesh. Unfortunately, this point cloud representation severely limits possible downstream applications.

In recent approaches~\cite{Corona21,Li22c}, a space of garments is learned with clothing items modeled as inflated SDFs and one single shared network to predict their deformations as a 3D displacement field. This makes deployment in real-world scenarios easier and allows the reconstruction of garments from images and 3D scans. However, the inflated SDF scheme reduces realism and precludes post-processing using  standard physics-based simulators or other cloth-specific downstream applications. Furthermore, both models are fully supervised and require a dataset of draped garments whose collection is extremely time-consuming.

Alleviating the need for costly ground-truth draped garments is tackled in~\cite{Santesteban22,Bertiche21b}, by introducing physics-based losses to train draping networks in a self-supervised manner. The approach of~\cite{Santesteban22} relies on a mass spring model to enforce the physical consistency of static garments deformed by different body poses. The method of~\cite{Bertiche21b} also accounts for variable body shapes and dynamic effects; furthermore, it incorporates a more realistic and expressive material model. Both methods, however, require training one network per garment, a limitation we remove. 



\section{Method}
\label{sec:method}

We aim to  realistically deform and drape generic garments over human bodies of various shapes and poses. To this end, we introduce the \algoname{} framework, presented in \cref{fig:framework}. It comprises a generative network shown on the left and a draping network shown on the right. Only the first is trained in a supervised manner, but using only static unposed garments meshes. This is key to avoiding having to run physics-based simulations to generate ground-truth data. Furthermore, we condition the draping network on latent vectors representing the input garments, which allows us to use the same network for very different garments, something that competing methods~\cite{Bertiche21b,Santesteban22} cannot do.

The generative network is a decoder trained using an encoder that turns a garment into a latent code $\bz$ that can then be decoded to an Unsigned Distance Function (UDF), from which a triangulated mesh can be extracted in a differentiable manner~\cite{Guillard22b}. The UDF representation allows us to accurately represent open surfaces and the many openings that garments typically feature. Since the top and bottom garments -- shirts and trousers -- have different patterns, we train one generative model for each. Both networks have the same architecture but different weights. 

The resulting {\it garment generative network} is only trained to output garments in a canonical shape, pose, and size that fit a neutral SMPL \cite{Loper15} body.  Draping the resulting garments to bodies in non-canonical poses is then entrusted to a {\it draping network}, again one for the top and one for the bottom. As in~\cite{Bertiche21b,Santesteban22,Li22c}, this network predicts vertex displacements w.r.t. the neutral position. The deformed garment is then skinned onto the articulated body model. To enable generalization to different tops and bottoms, we condition the draping process on the garment latent codes of the generative network, shown as $\bz_{\rm top}$ and $\bz_{\rm bot}$ in \cref{fig:framework}.

We use a small database of static unposed garments loosely aligned with bodies in the canonical position to train the two garment generating networks. This being done, we exploit physics-based constraints to train in a fully self-supervised manner the top and bottom draping networks for realism, without interpenetrations with the body and between the garments themselves.

\subsection{Garment Generative Network}
\label{sec:generative}

To encode garments into latent codes that can then be decoded into UDFs, we rely on a point cloud encoder that embeds points sampled from the unposed garment surface into a compact vector. This lets us obtain latent codes for previously unseen garments in a single inference pass from points sampled from its surface. This can be done given any arbitrary surface triangulation. Hence, it gives us the flexibility to operate on any given garment mesh.

We use DGCNN~\cite{Wang18b } as the encoder. It first propagates the features of points within the same local region at multiple scales and then aggregates them into a single global embedding by max pooling. We pair it with a decoder that takes as input a latent vector, along with a point in 3D space, and returns its (unsigned) distance to the garment. The decoder is a multi-layer perceptron (MLP) that relies on Conditional Batch Normalization \cite{DeVries17} for conditioning on the input latent vector. 

We train the encoder and the decoder by encouraging them to jointly predict distances that are small near the training garments' surface and large elsewhere. Because the algorithm we use to compute triangulated meshes from the predicted distances~\cite{Guillard22b} relies on the gradient vectors of the UDF field, we also want these gradients to be as accurate as possible~\cite{Atzmon20b,Zhao21a}. We therefore minimize the loss
\begin{equation} 
L_{garm} = L_{dist} + \lambda_g L_{grad} \; ,
\label{eq:gen_loss}
\end{equation} 
where $L_{dist}$ encodes our distance requirements, $ L_{grad}$ the gradient ones, and $\lambda_g$ is a weight balancing their influence.

More formally, at training time and given a mini-batch comprising $B$ garments, we sample a fixed number $P$ of points from the surface of each one. For each resulting point cloud $\mathbf{p}_i$ ($1 \! \leq \! i \! \leq \! B$),  we use the garment encoder $E_G$ to compute the latent code
\begin{equation}
    \bz_i = E_G(\mathbf{p}_i) \;
\end{equation}
and use it as input to the decoder $D_G$. It predicts an UDF field supervised with \cref{eq:gen_loss}, whose terms we define below. 

\textbf{Distance Loss.}
Having experimented with many different formulations of this loss, we found the following one both simple and effective. Given $N$ points $\{\mathbf{x}_{ij}\}_{j \leq N}$ sampled from the space surrounding the $i$-th garment, we pick a distance threshold $\delta$, clip all the ground-truth distance values $\{y_{ij}\}$ to it, and linearly normalize the clipped values to the range $[0,1]$. This yields normalized ground-truth values $\bar{y}_{ij}$ $=\min(y_{ij}, \delta)/\delta$. Similarly, we pass the output of the final layer of $D_G$ through a sigmoid function $\sigma(\cdot)$ to produce a prediction in the same range for point $\mathbf{x}_{ij}$
\begin{equation}
    \widetilde{y}_{ij} = \sigma(D_G(\mathbf{x}_{ij}, \bz_i)) \; .
\end{equation}
Finally, we take the loss to be
\begin{equation}
    \mathcal{L}_{dist} = \textsc{BCE}\left [ (\bar{y}_{ij})_{j \leq N}^{i \leq B} \; , \; (\widetilde{y}_{ij})_{j \leq N}^{i \leq B} \right ] ,
    \label{eq:gen_dist_loss}
\end{equation}
where $\textsc{BCE}[\cdot,\cdot]$ stands for binary cross-entropy. As observed in~\cite{Duan20}, the sampling strategy used for points $\mathbf{x}_{ij}$ strongly impacts training effectiveness. We describe ours in the supplementary. In our experiments, we set $\delta=0.1$, being the top and bottom garments normalized respectively into the upper and lower halves of the $[-1, 1]^3$ cube.

\textbf{Gradient Loss.}
Given the same sample points as before, we take the gradient loss to be
\begin{equation}
    \mathcal{L}_{grad} = \frac{1}{B N}\sum_{i,j} \left \| \mathbf{g}_{ij} - \widehat{\mathbf{g}_{ij}} \right \|_2^2 \;,
    \label{eq:gen_grad_loss}
\end{equation}
where $\mathbf{g}_{ij} = \nabla_{\mathbf{x}} y_{ij} \in \mathbb{R}^3$ is the ground-truth gradient of the $i$-th garment's UDF at $\mathbf{x}_{ij}$ and $\widehat{\mathbf{g}_{ij}} = \nabla_{\mathbf{x}} D_G(\mathbf{x}_{ij}, \mathbf{z}_i)$ the one of the predicted UDF, computed by backpropagation.

\subsection{Garment Draping Network}
\label{sec:draping}

We describe our approach to draping generic garments as opposed to specific ones and our self-supervised scheme. 
We assume that all garments are made of a single common fabric material, and we drape them in a quasi-static manner.

\subsubsection{Draping Generic Garments} 

We rely on SMPL \cite{Loper15} to parameterize the body in terms of shape ($\beta$) and pose ($\theta$) parameters. It uses Linear Blend Skinning  to deform a body template. Since garments generally follow the pose of the underlying body, we extend the SMPL skinning procedure to the 3D volume around the body for garment draping. Given a point $\bx \in \mathbb{R}^3$ in the garment space, its position $D(\bx, \beta, \theta, \bz)$ after draping becomes
\begin{align}
    D(\bx, \beta, \theta, \bz) &= W(\bx_{(\beta,\theta, \bz)}, \beta, \theta, \mathcal{W}(\bx)) \; , \label{eq:model} \\
    \bx_{(\beta, \theta, \bz)} &= \bx + \Delta x(\bx,\beta) + \Delta x_{\rm ref} (\bx, \beta, \theta, \bz) \; ,  \nonumber \\
    \Delta x_{\rm ref}(\bx, \beta, \theta, \bz) &=\mathcal{B}(\beta,\theta) \cdot \mathcal{M}(x, \bz) \; ,  \nonumber 
\end{align}
where $W(\cdot)$ is the SMPL skinning function, applied with blending weights $\mathcal{W}(\bx)$, over the point displaced by $\Delta x(\bx,\beta)$ and $\Delta x_{\rm ref}(\bx, \beta, \theta,  \bz)$. $\mathcal{W}(\bx)$ and $\Delta x(\bx,\beta)$ are computed as in \cite{Santesteban21,Li22c}. However, they only give an initial deformation for garments that roughly fits the underlying body. To refine it, we introduce a new term, $\Delta x_{\rm ref}(\bx, \beta, \theta,  \bz)$. It is a deformation field conditioned on body parameters $\beta$ and $\theta$, and on the garment latent code $\bz$ from the generative network. Following the linear decomposition of displacements in SMPL, it is the composition of an embedding $\mathcal{B}(\beta,\theta) \in \mathbb{R}^{N_\mathcal{B}}$ of body parameters and a displacement matrix $\mathcal{M}(x, \bz) \in \mathbb{R}^{N_\mathcal{B}\times 3}$ conditioned on $\bz$. Being conditioned on the  latent code $\bz$, $\Delta x_{\rm ref}$  can deform different garments differently, unlike the methods of~\cite{Bertiche21b,Santesteban22}. The number of vertices does not need to be fixed, since displacements are predicted separately for each vertex.

Since we have distinct encodings for the top and bottom garments, for each one we train two MLPs ($\mathcal{B}$, $\mathcal{M}$) to predict $\Delta x_{\rm ref}$. The other MLPs for $\mathcal{W}(\cdot)$ and $\Delta x(\cdot)$ are shared.

\subsubsection{Self-Supervised Training} 

We first learn the weights of $\mathcal{W}(\cdot)$ and $\Delta x(\cdot)$ as in \cite{Santesteban21,Li22c}, which does not require any annotation or simulation data but only the blending weights and shape displacements of SMPL. We then train our deformation fields $\Delta x_{\rm ref}$ in a fully self-supervised fashion by minimizing the physics-based losses introduced below. In this way, we completely eliminate the huge cost that extensive simulations would entail. 

\textbf{Top Garments.} For upper body garments -- shirts, t-shirts, vests, tank tops, etc. -- the deformation field is trained using the loss from \cite{Santesteban22}, expressed as
\begin{equation}
    \mathcal{L}_{top} = \mathcal{L}_{strain} + \mathcal{L}_{bend} + \mathcal{L}_{gravity} + \mathcal{L}_{col} \; , \label{eq:physics}
\end{equation}
where $\mathcal{L}_{strain}$ is the membrane strain energy of the deformed garment, $\mathcal{L}_{bend}$ the bending energy caused by the folding of adjacent faces, $\mathcal{L}_{gravity}$ the gravitational potential energy, and $\mathcal{L}_{col}$ a penalty for collisions between body and garment. Unlike in~\cite{Santesteban22}, we only consider the quasi-static state after draping, that is, without acceleration.

\textbf{Bottom Garments.} Due to gravity, bottom garments, such as trousers, would drop onto the floors if we used only the loss terms of \cref{eq:physics}. We thus introduce an extra loss term to constrain the deformation of vertices around the waist and hips. The loss becomes
\begin{align}
    \mathcal{L}_{bottom} &= \mathcal{L}_{strain} + \mathcal{L}_{bend} + \mathcal{L}_{gravity} + \mathcal{L}_{col} + \mathcal{L}_{pin},   \nonumber \\ 
     \mathcal{L}_{pin}     &= \sum_{v\in V} |\Delta x_y|^2+\lambda(|\Delta x_x|^2+|\Delta x_z|^2) \; , \label{eq:pin} 
\end{align}
where $V$ is the set of garment vertices whose closest body vertices are located in the region of the waist and hips. See supplementary material for details. The terms $\Delta x_x$, $\Delta x_y$ and $\Delta x_z$ are the deformations along the X, Y and Z axes, respectively. $\lambda$ is a positive value smaller than 1 that penalizes deformations along the vertical direction (Y axis) and produces natural deformations along the other directions.

\textbf{Top-Bottom Intersection.} To ensure that the top and bottom garments do not intersect with each other when we drape them on the same body, we define a loss $\mathcal{L}_{IS}$ that ensures that when the top and the bottom garments overlap, the bottom garment vertices are closer to the body mesh than the top ones, which prevents them from intersecting
-- this is arbitrary, and the following could be formulated the other way around.
To this end, we introduce an Intersection Solver (IS) network. It predicts a displacement correction $\Delta x_{IS}$, added only when draping bottom garments as
\begin{equation}
   \tilde{\bx}_{(\bz_{top}, \bz_{bot})} = \bx_{(\bz_{bot})} + \Delta x_{IS}(\bx, \bz_{top}, \bz_{bot}) \; ,
\end{equation}
where we omit the dependency of $\tilde{\bx}$, $\bx$ and $\Delta x_{IS}$ on the body parameters $(\beta, \theta)$ for simplicity. $\bz_{top}$ and $\bz_{bot}$ are the latent codes of the top and bottom garments, and $\bx_{(\bz_{bot})}$ is the input point displaced according to \cref{eq:model}. The skinning function of \cref{eq:model} is then applied to $\tilde{\bx}_{(\bz_{top}, \bz_{bot})}$ for draping.
$\Delta x_{IS}(\cdot)$ is implemented as a simple MLP and trained with
\begin{equation}
    \mathcal{L}_{IS} = \mathcal{L}_{bottom} + \mathcal{L}_{layer}, \label{eq:intersection}
\end{equation}
where $\mathcal{L}_{layer}$ is a loss whose minimization requires the top and bottom garments to be separated from each other. We formulate it as 
\begin{equation}
    \mathcal{L}_{layer} = \; \sum_{\mathclap{v_{B}\in C}} \; max(d_{bot}(v_{B})-\gamma d_{top}(v_{B}), 0) \;,
\end{equation}
where $C$ is the set of body vertices covered by both the top and bottom garments, $d_{top}(\cdot)$ and $d_{bot}(\cdot)$ the distance to the top and the bottom garments respectively, and $\gamma$ a positive value smaller than 1 (more details in the supplementary).


\section{Experiments}
\label{sec:experiments}


\begin{figure}
    \centering
    \includegraphics[width=0.45\textwidth]{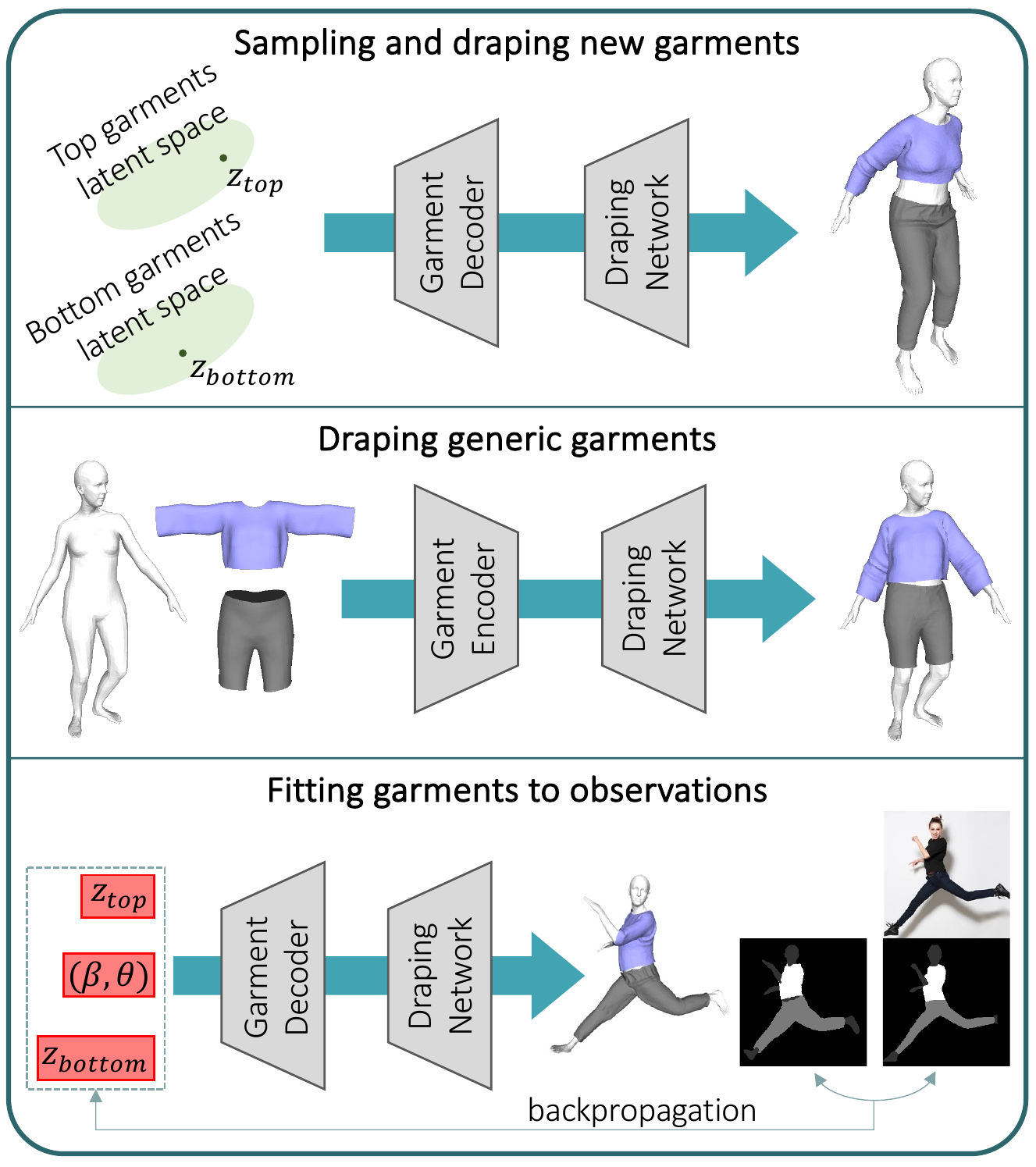}
    \vspace{-0.25cm}
    \caption{\textbf{Overview of \algoname{} applications.} \textbf{Top:} New garments can be sampled from the latent spaces of the generative networks, and deformed by the draping networks to fit to a given body. \textbf{Center:} The garment encoders and the draping networks form a general purpose framework to drape any garment with a single forward pass.  \textbf{Bottom:} Being a  differentiable parametric model, our framework can reconstruct 3D garments by fitting observations such as images or 3D scans. The red boxes indicate the parameters optimized in this process.}
    \label{fig:applications}
    \vspace{-0.35cm}
\end{figure}

We first describe our experimental setup and test \algoname{} for the different purposes depicted by \cref{fig:applications}. They include reconstructing different kinds of garments and editing them by manipulating their latent codes. We then gauge the draping network both qualitatively and quantitatively. Finally, we use \algoname{} to reconstruct garments from images and 3D scans.

\subsection{Settings, Datasets and Metrics}
\label{sec:settings}

\textbf{Datasets.}
Both our generative and draping networks are trained with garments from CLOTH3D~\cite{Bertiche20}, a synthetic dataset that contains over 7K sequences of animated 3D humans parametrized used the SMPL model and wearing different garments. Each sequence comprises up to 300 frames and features garments coming from different templates. For training, we randomly selected 600 top garments (t-shirts, shirts, tank tops, etc.) and 300 bottom garments (both long and short trousers). Neither for the generative nor for the draping networks did we use the simulated deformations of the selected garments. Instead, we trained the networks using only garment meshes on average body shapes in T-pose. By contrast, for testing purposes, we selected random clothing items -- 30 for top garments and 30 bottom ones -- and considered {\it whole} simulated sequences.

\textbf{Training.}
We train two different models for top and bottom garments, both for the generative and for the draping parts of our framework. First, the generative models are trained on the 600/300 neutral garments. Then, with the generative networks weights frozen, we train the draping networks by following~\cite{Santesteban22}: body poses $\theta$ are sampled randomly from the AMASS~\cite{mahmood19} dataset, and shapes $\beta$ uniformly from $[-3,3]^{10}$ at each step. The other hyperparameters are given in the supplementary material.

\textbf{Metrics.}
We report the Euclidean distance (ED), interpenetration ratio between body and garment (B2G), and intersection between top and bottom garments (G2G).
ED is computed between corresponding vertices of the considered meshes. B2G is the area ratio between the garment faces inside the body and the whole surface as in~\cite{Li22c}. Since CLOTH3D exclusively features pairs of top/bottom garments with the bottom one closer to the body, G2G is computed by detecting faces of the bottom garment that are outside of the top one, and taking the area ratio between those and the overall bottom garment surface.

\subsection{Garment Paramerization}
\label{sec:experiments_generative}

We first test the encoding-decoding scheme of \cref{sec:generative}. 

\begin{figure}
    \centering
    \includegraphics[width=0.38\textwidth]{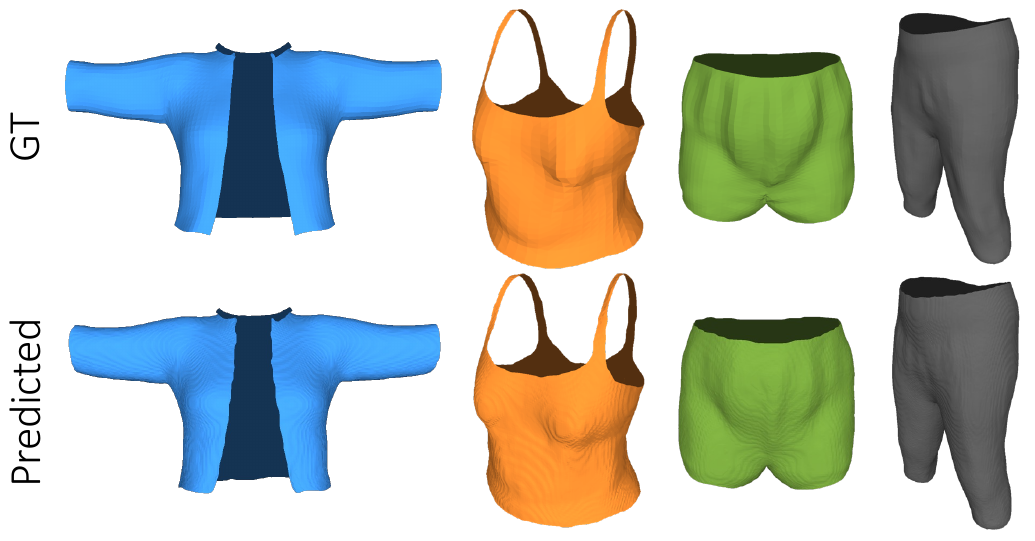}
    \vspace{-0.35cm}
    \caption{\textbf{Generative network: reconstruction of unseen garments in neutral pose/shape.} The latent codes are obtained with the garment encoder, then decoded into open surface meshes.
    }
    \label{fig:generative_rec}
    \vspace{-0.45cm}
\end{figure}

\textbf{Encoding-Decoding Previously Unseen Garments.}
The generative network of  \cref{fig:framework}  is designed to project garments into a latent space and to reconstruct them from the resulting latent vectors. In \cref{fig:generative_rec}, we visualize reconstructed previously-unseen garments from CLOTH3D. The reconstructions are faithful to the input garments, including fine-grained details such as the shirt collar on the left or the shoulder straps of the tank top.

\begin{figure}
    \centering
    \includegraphics[width=0.45\textwidth]{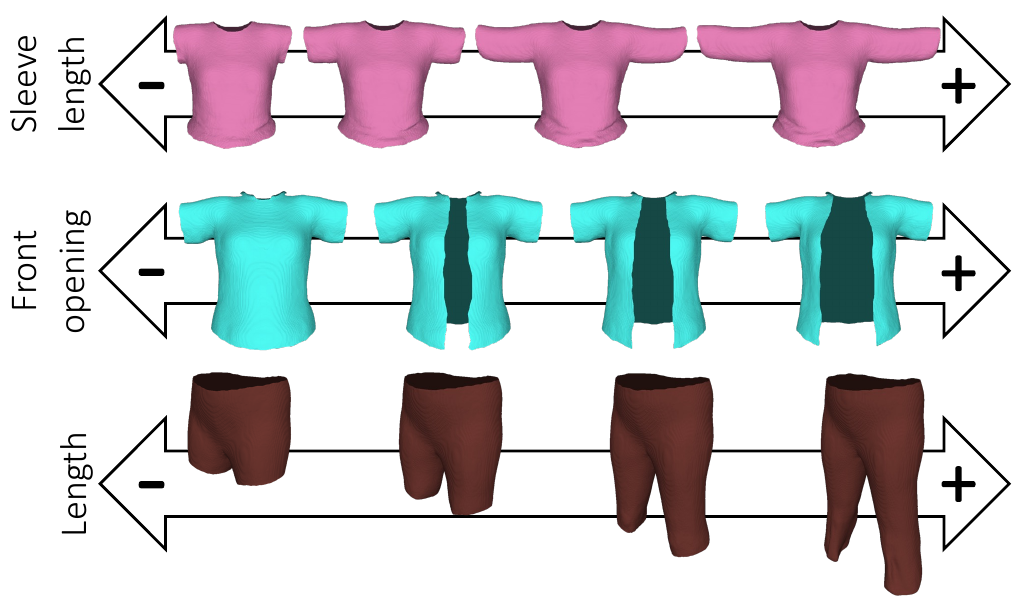}
    \vspace{-0.45cm}
    \caption{\textbf{Garment editing.} The latent codes produced by the garment encoder can be manipulated to edit specific features of the corresponding garments, without altering the overall geometry.}
    \label{fig:latent_manip}
    \vspace{-0.2cm}
\end{figure}

\textbf{Semantic Manipulation of Latent Codes.}
Our framework enables us to edit a garment by manipulating its latent code. For the resulting edits to have a semantic meaning, we assigned binary labels corresponding to features of interest to 100 training garments. For instance, we labeled garments as having ``short sleeves'' (label = 0) or ``long sleeves'' (label = 1). Then, we fit a linear logistic regressor to the garment latent codes.  After training, the regressor weights indicate which dimensions of the latent space control the feature of interest. To this end, we first apply min-max normalization to the absolute weight values and then zero out the ones below a certain threshold, empirically set to 0.5. The remaining non-zero weights indicate which dimensions of the latent codes should be increased or decreased to edit the studied feature.  To create \cref{fig:latent_manip}, we applied this simple procedure to control the sleeve length and the front opening for top garments along with the length for bottom garments. As can be seen from the figure, our latent representations give us the ability to edit a specific garment feature while leaving other aspects of the garment geometry unchanged.


\subsection{Garment Draping}
\label{sec:experiments_draping}

We now turn to the evaluation of the draping network and compare its performance to those of DeePSD~\cite{Bertiche21} or DIG~\cite{Li22c}, two \textit{fully supervised} learning methods trained on CLOTH3D.  DeePSD takes the point cloud of the garment mesh as input and predicts blending weights and pose displacements for each point; DIG drapes garments with a learned skinning field that can be applied to generic 3D points, but is similar for all garments. We chose those because, like \algoname{}, they both can deform garments of arbitrary geometry and topology. 


\begin{figure*}
    \centering
    \includegraphics[width=0.87\textwidth]{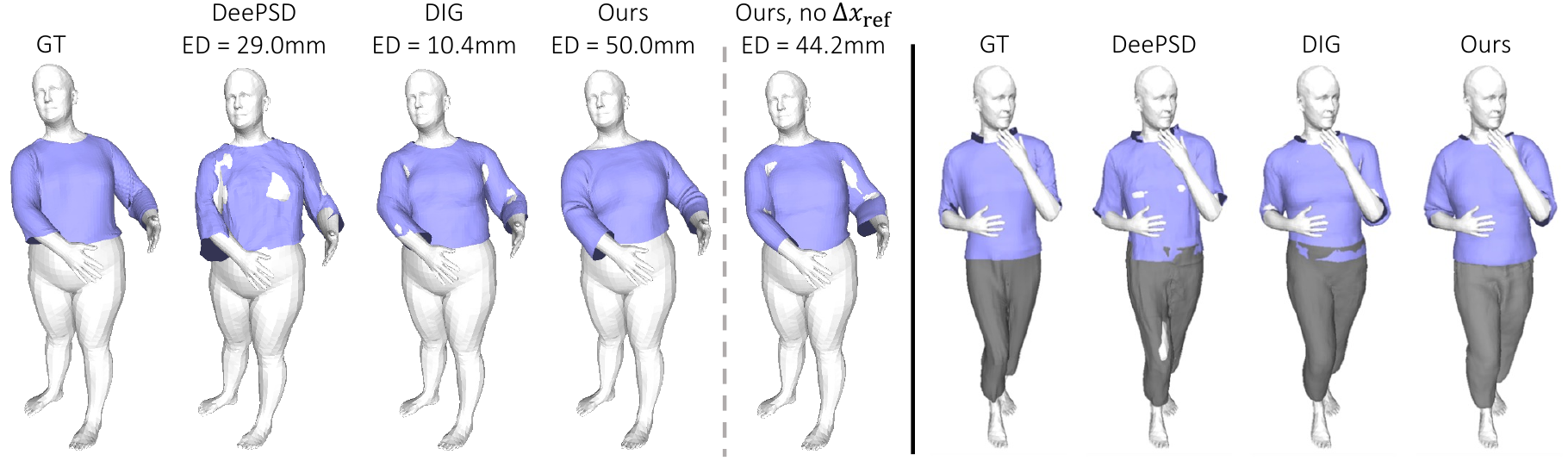}
    \vspace{-0.35cm}
    \caption{\textbf{Comparison between DeePSD, DIG and our method.} Ours is more realistic despite having the highest Euclidean distance (ED) error (\textbf{left}), and has less intersection between garments (\textbf{right}).
    \textbf{Left} also shows that $\Delta x_{\rm ref}$ is necessary for realistic deformations.
    }
    \vspace{-0.45cm}
    \label{fig:qualitative}
\end{figure*}


\begin{table}
  \begin{center}
    \scalebox{.8}{
          \begin{tabular}{c | c | c | c}
          \toprule
            & DeePSD & DIG & Ours \\
           \midrule
           ED-top (mm)  & 28.1 & 29.6 & 47.9 \\
           ED-bottom (mm) & 18.3 & 20.0 & 27.3 \\
           \midrule
           B2G-top (\%) $\downarrow$    & 7.2 & 1.8 & \textbf{0.9}\\
           B2G-bottom (\%) $\downarrow$ & 3.4 & 0.8 & \textbf{0.3}  \\
           G2G (\%) $\downarrow$   & 2.0  & 4.0 & \textbf{0.5} \\
          \bottomrule
          \end{tabular}
          }
    \end{center}
    \vspace{-0.5cm}
    \caption{\textbf{Draping unseen garment meshes.} Comparison between DeePSD, DIG and our method, for top and bottom garments: Euclidean distance (ED), intersections with the body (B2G) and between garments (G2G) as ratio of intersection areas.}
    \label{tab:draping}
    \vspace{-0.5cm}
\end{table}


\begin{figure}
    \centering
    \includegraphics[width=0.45\textwidth]{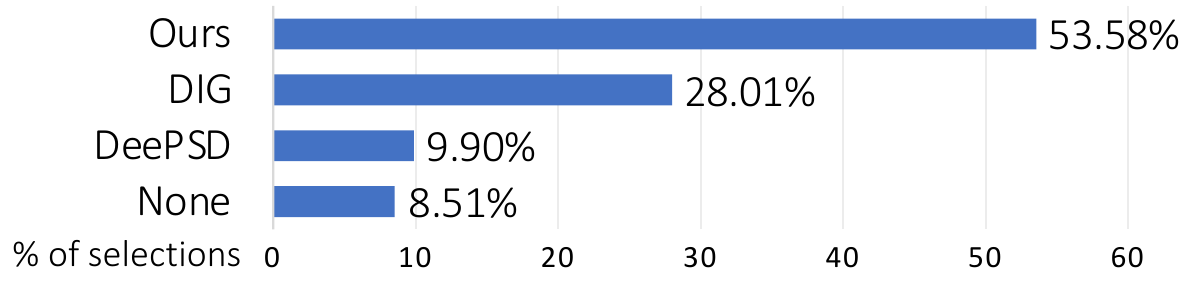}
    \vspace{-0.25cm}
    \caption{\textbf{Human evaluation of draping results.} When shown draping results of our method, DIG and DeePSD, evaluators selected ours as the most realistic one in more than half of the cases. \textit{None} refers to the case when they had no clear preference.
    }
    \label{fig:human_eval}
    \vspace{-0.45cm}
\end{figure}

\textbf{Draping Unseen Meshes.}
We drape previously unseen garments on different bodies in random poses. We first encode the garments and use the resulting latent codes to condition the draping network, whose inference takes $\sim$5ms.

We provide qualitative results in \cref{fig:qualitative} and report quantitative ones in \cref{tab:draping}. Despite being completely self-supervised, \algoname{} delivers the lowest ratio of body-garment interpenetrations (B2G) for both top and bottom garments and the least intersections between them (G2G).

However, \algoname{} also yields higher ED values, which makes sense because there is more than one way to satisfy the physical constraints and to achieve realism. Hence, in the absence of explicit supervision, there is no reason for the answer picked by \algoname{} to be exactly the same as the one picked by the simulator. In fact, as argued in~\cite{Bertiche21b} and illustrated by \cref{fig:qualitative}, which is representative in terms of ED, a low ED value does not necessarily correspond to a realistic draping. To confirm this, we conducted a human evaluation study by sharing a link to a website on friends groupchats. We gave no further instructions or details besides those given on the site and reproduced in the supplementary material. The website displays 3 drapings of the same garment over the same posed body, one computed using our method and the others using the other two. The users were asked to select which one of the three seemed more realistic and more pleasant, with a fourth potential response being ``none of them". We obtained feedback from 187 different people. A total of 1258 individual examples were rated and we collected 3738 user opinions. In other words, each user expressed 20 opinions on average. The chart in \cref{fig:human_eval} shows that our method was selected more than 50\% of the times, with a large gap over the second best, DIG~\cite{Li22c}, selected less than 30\% of the time. This result confirms that \algoname{} can drape garments with better perceptual quality than the competing methods.


\begin{figure}
    \centering
    \includegraphics[width=0.46\textwidth]{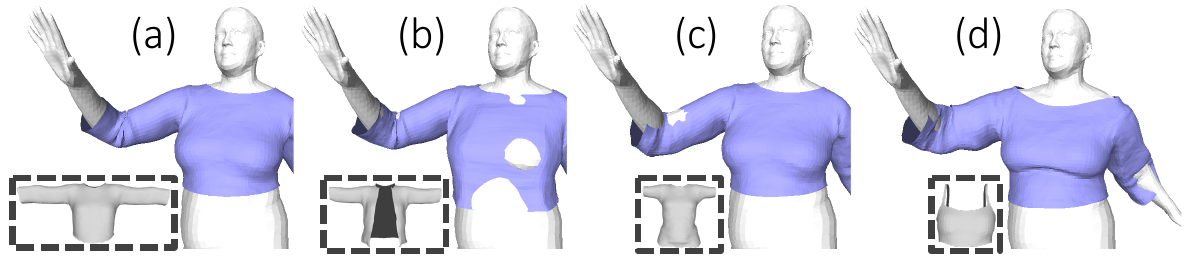}
    \vspace{-0.25cm}
    \caption{\textbf{Switching input latent codes of the draping network.} Draping the same shirt by conditioning the draping network with \textbf{(a)} the corresponding latent code, \textbf{(b)} the code of an open vest, \textbf{(c)} of a t-shirt and \textbf{(d)} of a tank top. Gray meshes in dashed boxed are the garments corresponding to the input latent codes.}
    \label{fig:switch}
    \vspace{-0.35cm}
\end{figure}

\textbf{Ablation Study.}
In \cref{fig:switch}, we show what happens when the draping network is conditioned with a latent code of a garment that does not match the input one. This creates unnatural deformations on the front when using the code of a shirt with a front opening to deform a shirt without an opening. Similarly, the sleeves penetrate the arms when conditioning with the code of a short sleeves shirt. This demonstrates that the draping network truly exploits the latent codes to predict garment-dependent deformation fields.

In \cref{fig:qualitative} \textbf{left} we show that removing our novel displacement term $\Delta x_{\rm ref}(\cdot)$ from \cref{eq:model} leads to unrealistic results.

We also ablate the influence of our Intersection Solver and observe that G2G increases from 0.5\% to 1.1\% without it. This demonstrates the effectiveness of this component at reducing collisions between top and bottom garments.

\subsection{Fitting Observations}
\label{sec:fitting}

Since our method is end-to-end differentiable, it can be used to reconstruct 3D models of people and their garments from partial observations, such as 2D images and 3D scans.

\textbf{Fitting Images.} 
Given an image of a clothed person, we use the algorithm of~\cite{Rong20,Yang20f} to get initial estimates for the body parameters $(\beta,\theta)$ and a segmentation mask $\textbf{S}$. Then, starting with the mean of the learned codes $\bz$, we reconstruct a mesh for the body and its garments by minimizing
\begin{equation}
\scalebox{0.9}{$
\begin{aligned}
  L(\beta,\theta,\bz) &= L_{\text{IoU}}(R(D(\textbf{G},\beta,\theta, \bz), \text{SMPL}(\beta,\theta)), ~\textbf{S})\; , \label{eq:fit_image} \\
  \textbf{G} &= \text{MeshUDF}(D_G(\bz)) \; ,
\end{aligned}
$}
\end{equation}
w.r.t. $\bz$, $\beta$ and $\theta$, where $L_{\text{IoU}}$ is the IoU loss \cite{Li21g} in pixel space penalizing discrepancies between 2D masks, $R(\cdot)$ is a differentiable mesh renderer \cite{Pytorch3D}, and $\textbf{G}$ is the set of vertices of the garment mesh reconstructed with our garment decoder using $\bz$. $D(\cdot)$ and $\text{SMPL}(\cdot)$ are the garment and body skinning functions defined in \cref{eq:model} and in \cite{Loper15}, respectively. To ensure pose plausibility, $\theta$ is constrained by an adversarial pose prior \cite{Davydov22a}. 

\begin{figure}
    \centering
    \includegraphics[width=0.47\textwidth]{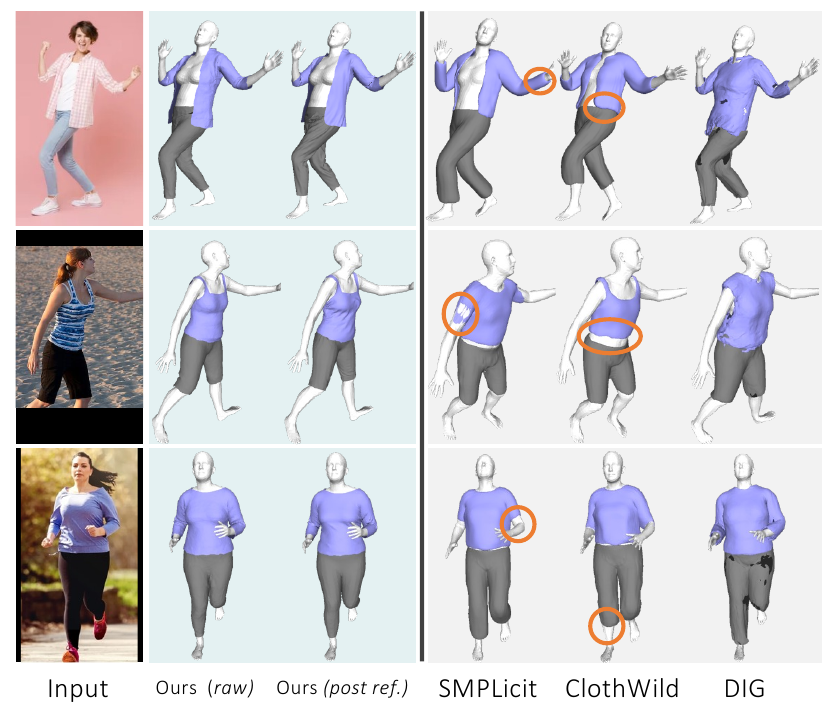}
    \vspace{-0.15cm}
    \caption{\textbf{Recovering garments and bodies from images.} From left to right we show the input image and the 3D models recovered with our method (without and with post-refinement), and competitors methods: SMPLicit~\cite{Corona21}, ClothWild~\cite{Moon22}, DIG~\cite{Li22c}.}
    \label{fig:image}
    \vspace{-0.35cm}
\end{figure}

For the sake of simplicity, \cref{eq:fit_image} formulates the reconstruction of a single garment $\textbf{G}$. In practice, we extend this formulation to both the top and the bottom garments shown in the target image.
\cref{fig:image} depicts the results of minimizing this loss.  It outperforms the state-of-the-art methods SMPLicit~\cite{Corona21}, ClothWild~\cite{Moon22} and DIG~\cite{Li22c}. The garments we recover follow the ones in the input image with higher fidelity and visual quality, without  interpenetration between the body and the garments or between the two garments.

After this optimization, we can further refine the result by minimizing the physics-based objectives of \cref{eq:physics} w.r.t. the per-vertex displacements of the reconstructed garments, as opposed to w.r.t. the latent vectors. We describe this procedure in the supplementary material. As shown in the third column of \cref{fig:image}, this further boosts the realism of the reconstructed garments. Note that this refinement is feasible thanks to the open surface representation allowed by our UDF model. Applying these physically inspired losses to an inflated garment, as produced by SMPLicit, ClothWild and DIG, yields poor results with many self-intersections, as shown in the supplementary material.


\begin{figure}
    \centering
    \includegraphics[width=0.4\textwidth]{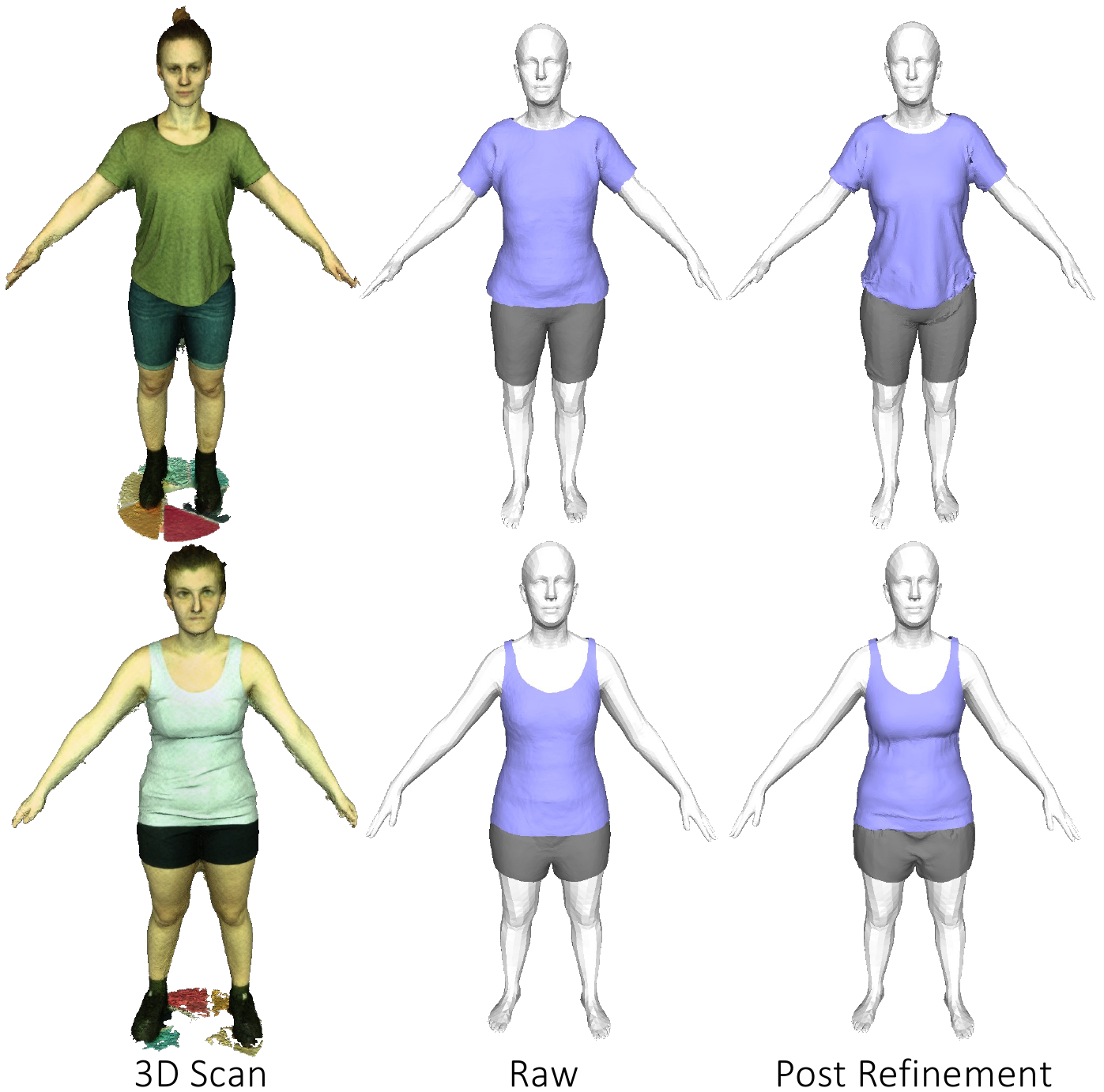}
    \vspace{-0.2cm}
    \caption{\textbf{Recovering garments and bodies from 3D scans.} We show 3D models recovered with our method from scans of the SIZER dataset~\cite{Tiwari20}. \textit{Raw} indicates the model recovered with \cref{eq:fit_scan} from the 3D scan. \textit{Post Refinement} refers to the models further refined with the physics-based losses.}
    \label{fig:scan}
    \vspace{-0.35cm}
\end{figure}

\textbf{Fitting 3D scans.} 
Given a 3D scan of a clothed person and segmentation information, we apply a strategy similar to the one presented above and minimize
\begin{equation}
  \scalebox{0.9}{$
  L(\beta,\theta,\bz) = d(D(\textbf{G},\beta,\theta, \bz), ~\textbf{S}_\textbf{G}) + \vec{d}(\text{SMPL}(\beta,\theta), ~\textbf{S}_\textbf{B}), \label{eq:fit_scan}
  $}
\end{equation}
w.r.t. $\bz$, $\beta$ and $\theta$, where $\textbf{S}_\textbf{G}$ and $\textbf{S}_\textbf{B}$ denote the segmented garment and body scan points, and $d(a,b)$ and $\vec{d}(a,b)$ are the bidirectional and the one-directional Chamfer distance from $b$ to $a$.
Similarly to \cref{eq:fit_image}, we apply \cref{eq:fit_scan} to recover both the top and bottom garments.
\cref{fig:scan} shows our fitting results for some scans of the SIZER dataset \cite{Tiwari20}.
The recovered 3D models closely match the input scans. Moreover, we can also apply a post-refinement procedure similar to the one described above, by minimizing both the physics-based losses from \cref{eq:physics} and the Chamfer distance to the input scan w.r.t. the 3D coordinates of the vertices of the reconstructed models. This leads to even more realistic results, with fine wrinkles aligning to the input scans.


\section{Conclusion}
\label{sec:conclusion}

We have shown that physics-based self-supervision can be leveraged to learn a single parameterization for many different garments to be draped on human bodies in arbitrary poses. Our approach relies on UDFs to represent garment surfaces and on a displacement field to drape them, which enables us to handle a continuous manifold of garments without restrictions on their topology. Our whole pipeline is differentiable, which makes it suitable for solving inverse problems and for modeling clothed people from image data. 

Future work will focus on modeling dynamic poses instead of only static ones. This is of particular relevance for loose clothes, where our reliance on the SMPL skinning prior should be relaxed. Moreover, we will investigate replacing our current global latent code by  a set of local ones to yield finer-grained control both for garment editing and draping.

\clearpage
\clearpage
\textbf{Acknowledgement.} This project was supported in part by the Swiss National Science Foundation.
{\small
\bibliographystyle{ieee_fullname}
\bibliography{string,geom,graphics,learning,vision,misc}
}
\clearpage
\clearpage

\section*{Supplementary Material}

In this appendix we first provide more details about our networks and their architectures in \cref{sec:supp_net_hyperparam}.

In \cref{sec:supp_losses} we expand on the choice and formulations of some loss terms we use.
Importantly, in \cref{subsec:supp_refinement} we explain the physics-based refinement procedure used in the main paper, and show that \textit{modelling garments as open surfaces is necessary for it.}

Then in \cref{sec:supp_results} we report additional quantitative and qualitative results of our pipeline and the runtime of its components. Finally, in \cref{sec:supp_human_eval} we describe how human ratings were collected.

\section{Network Architectures and Training}
\label{sec:supp_net_hyperparam}
\subsection{Garment Generative Network}


\begin{figure*}
    \centering
    \includegraphics[width=0.95\textwidth]{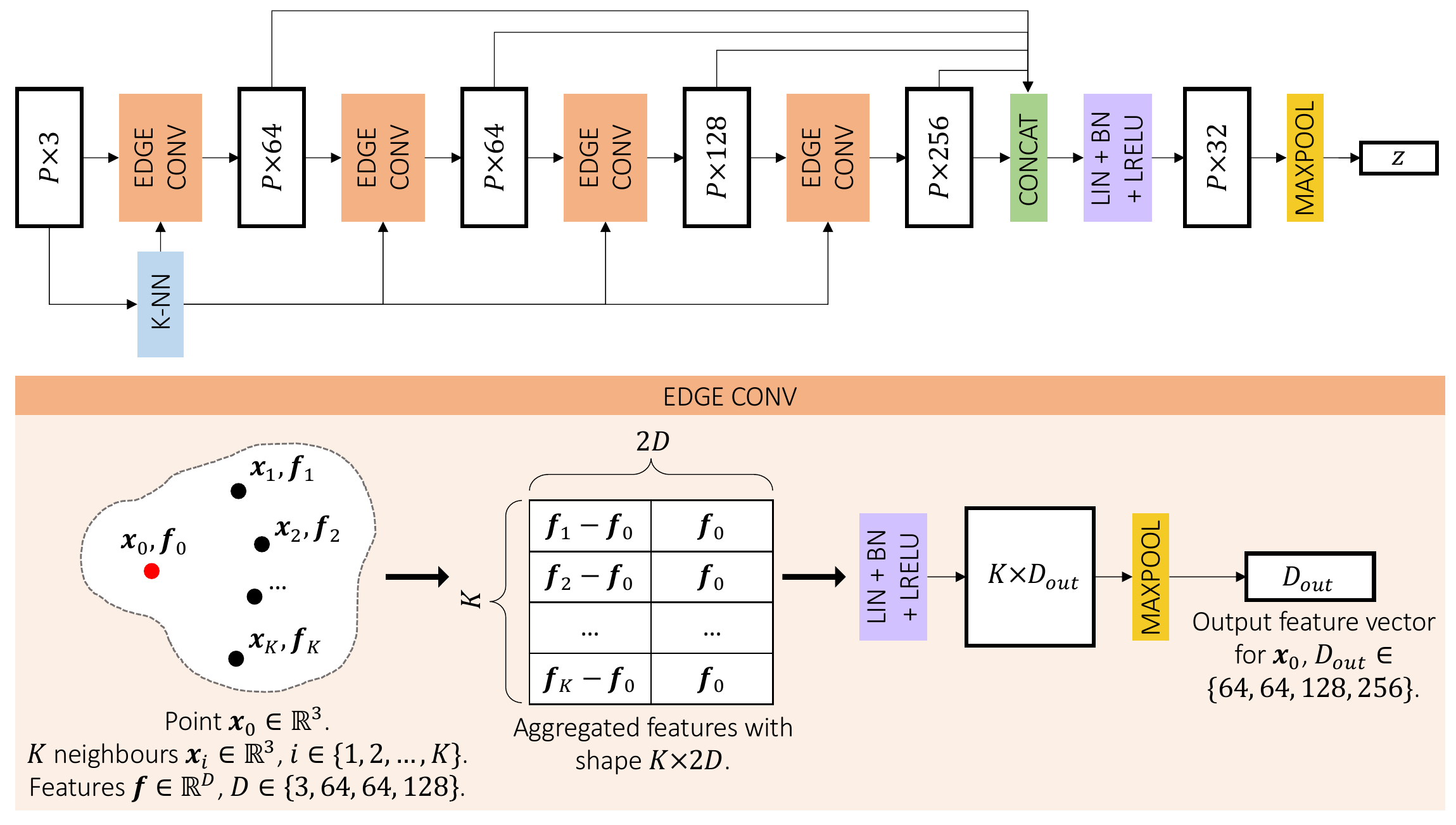}
    \caption{\textbf{DGCNN point cloud encoder.} We adopt DGCNN~\cite{Wang18b} as the point cloud encoder of our garment generative network. The input cloud is passed through four \textit{edge convolutions}, which gather features of local neighborhoods of points to project them into higher dimensional spaces. The features from all the layers are then concatenated and projected to the final desired dimension. Max pooling is finally used to obtain the latent code $\mathbf{z}$ for the input cloud. \textit{CONCAT} stands for features concatenation, while \textit{LIN + BN + LRELU} represents a linear layer followed by batch normalization and leaky ReLU.}
    \label{fig:dgcnn}
\end{figure*}

\subsubsection{Garment Encoder}
To encode a given garment into a compact latent code, we first sample $P$ points from its surface and then we feed them to a DGCNN~\cite{Wang18b} encoder, detailed in \cref{fig:dgcnn}. The input point cloud is processed by four \textit{edge convolution} layers, which project the input 3D points into features with increasing dimensionality -- \ie{}, 64, 64, 128 and finally 256.

Each edge convolution layer works as follows. For each input point, the features from its $K$ neighbours are collected and used to prepare a matrix with $K$ rows. Each row is the concatenation of two vectors: $\mathbf{f}_i - \mathbf{f}_0$ and $\mathbf{f}_0$, $\mathbf{f}_i$ and $\mathbf{f}_0$ being respectively the feature vector of the $i$-th neighbour and the feature vector of the considered point. Each row of the resulting matrix is then transformed independently to the desired output dimension. The output feature vector for the considered point is finally obtained by applying max pooling along the rows of the produced matrix.

The original DGCNN implementation recomputes the neighborhoods in each edge convolution layer, using the distance between the feature vectors as metric. This can be explained by the original purposes of DGCNN, \ie{}, point cloud classification and part segmentation. Since we are interested in encoding the geometric details of the input point cloud, we compute neighborhoods only once based on the euclidean distance of the points in the 3D space and reuse this information in every edge convolution layer. We set $K=16$ in our experiments.

The feature vectors from the four edge convolutions are then concatenated to form a single vector with 512 elements, that is fed to a final linear layer paired with batch normalization and leaky ReLU. Such layer projects the 512 sized vectors into the final desired dimension, which is 32 in our case.
The final latent code is obtained by compressing the feature matrix with shape $P \times 32$ along the first dimension with max pooling.


\begin{figure*}
    \centering
    \includegraphics[width=0.95\textwidth]{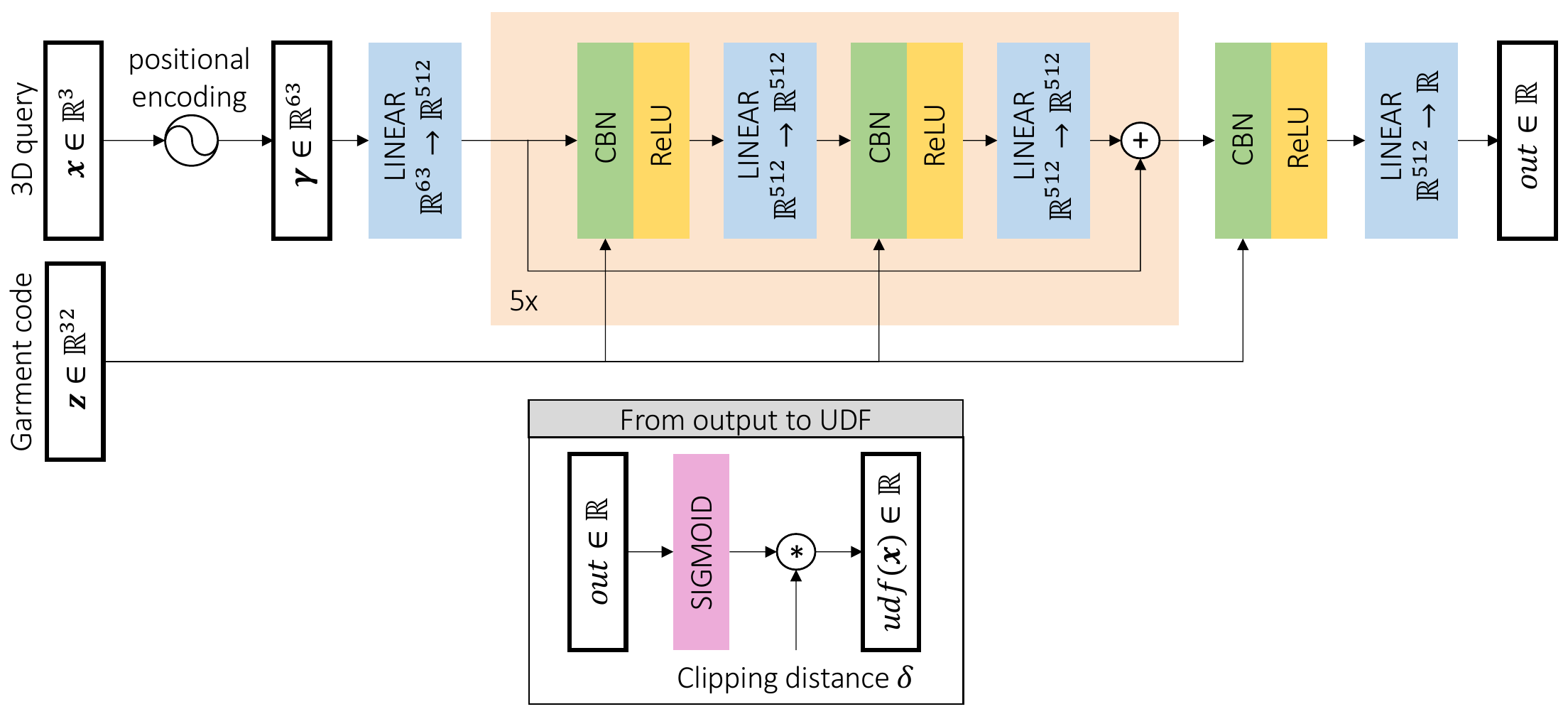}
    \caption{\textbf{UDF decoder.} Given a 3D query and a garment latent code, the decoder of our garment generative network is trained to predict the UDF of the input query w.r.t. the surface of the garment. The latent code is used to condition the prediction by the means of Conditional Batch Normalization (CBN) \cite{DeVries17}. Since we trained the decoder with the binary cross-entropy loss, its outputs need to be converted to UDF values by applying the sigmoid function and then scaling the result with the UDF clipping distance $\delta$.}
    \label{fig:cbndec}
\end{figure*}

\subsubsection{Garment Decoder}
The garment generative network features an implicit decoder that can predict the unsigned distance field of a garment starting from its latent code. More specifically, the decoder is a coordinate-based MLP that takes as inputs the garment latent code and a 3D query. Using the latent code as condition, the decoder predicts the unsigned distance from the query to the garment surface.

Our UDF decoder, shown in \cref{fig:cbndec}, is inspired by \cite{Mescheder19}. The input 3D query is first mapped to a higher dimensional space ($\mathbb{R}^{63}$) with the positional encoding proposed in \cite{Mildenhall20}, which is known to improve the capability of the network to approximate high frequency functions. The encoded query is then mapped with a linear layer to $\mathbb{R}^{512}$ and then goes through 5 residual blocks. 
The output of each block is computed as $\mathbf{f}_{out} = \mathbf{f}_{in} + \Delta \mathbf{f}$, where $\mathbf{f}_{in}$ is the input vector and $\Delta \mathbf{f}$ is a residual term predicted by two consecutive linear layers starting from $\mathbf{f}_{in}$. The size of the feature vector is 512 across the whole sequence of residual blocks.
The output of the last block is mapped to the scalar output $out \in \mathbb{R}$ with a final linear layer.

All the linear layers but the output one are paired with Conditional Batch Normalization (CBN) \cite{DeVries17} and ReLU activation function. CBN is used to condition the MLP with the input latent code $\mathbf{z}$. In more details, each CBN module applies standard batch normalization \cite{Ioffe15} to the input vectors, with the difference that the parameters of the final affine transformation are not learned during the training but are instead predicted at each inference step by dedicated linear layers starting from $\mathbf{z}$.

Finally, recall that our generative network is trained with the binary cross-entropy loss. Thus, the output of the decoder must be converted to the corresponding UDF value by first applying the sigmoid function and then scaling the result with the UDF clipping distance $\delta$, which we set to 0.1 in our experiments. Such procedure is indeed the dual of the one applied on the UDF ground-truth labels during training to normalize them in the range $[0,1]$.

\subsubsection{Surface Sampling} 
We sample supervision points with a probability inversely proportional to the distance to the surface: $30\%$ of the points are sampled directly on the input surface, $30\%$ are sampled by adding gaussian noise with $\epsilon$ variance to surface points, $30\%$ are obtained with gaussian noise with $3\epsilon$ variance, and the remaining ones are gathered by sampling uniformly the bounding box in which the garment is contained. Since in our experiments, the top and bottom garments are normalized respectively into the upper and lower halves of the $[-1, 1]^3$ cube, we set $\epsilon=0.003$.

\subsection{Draping Network}
The networks $\mathcal{W}(\bx) \in \mathbb{R}^{24}$ and $\Delta x(\bx,\beta) \in \mathbb{R}^{3}$ that predict blending weights and coarse displacements are implemented by a 9-layer multilayer perceptron (MLP) with a skip connection from the input layer to the middle. Each layer has 256 nodes except the middle and the last ones. ReLU is used as the activation function. The body-parameter-embedding module $\mathcal{B}(\beta,\theta) \in \mathbb{R}^{128}$ and the displacement-matrix module $\mathcal{M}(x, \bz) \in \mathbb{R}^{128\times 3}$ for $\Delta x_{\text{ref}}$ are implemented by a 5-layer MLP with LeakyReLU activation in-between. Each layer has 512 nodes except the last one. $\Delta x_{IS}$ uses the same architecture as $\Delta x_{\text{ref}}$.

\subsection{Training Hyperparameters}
The generative models (top/bottom ones) are trained on the 600/300 neutral garments for 4000 epochs, using mini-batches of size $B=4$. Each item of the mini-batch contains an input point cloud with $P=10,000$ points and $N=20,000$ random UDF 3D queries. The dimension of the latent codes is set to 32 for both top and bottom garments, and we set $\lambda_g = 0.1$ in 
\begin{equation} 
   \mathcal{L}_{garm} = \mathcal{L}_{dist} + \lambda_g \mathcal{L}_{grad} \; .
   \label{eq:supp_gen_loss}
\end{equation} 

The draping networks are trained for 250K iterations with mini-batches of size 18, where each item is composed of the vertices of one garment paired with one body shape and pose. We set $\lambda=0.1$ for $\mathcal{L}_{pin}$ and $\gamma=0.5$ for $\mathcal{L}_{layer}$.

Both the generative and the draping networks are trained with Adam optimizer \cite{Kingma15} and learning rates set to $0.0001$ and $0.001$ respectively.

\section{Loss Terms and Ablation Studies}
\label{sec:supp_losses}
\subsection{\texorpdfstring{$\mathcal{L}_{garm}$}{Lgarm} for Garment Reconstruction}


\begin{figure*}
    \centering
    \includegraphics[width=0.99\textwidth]{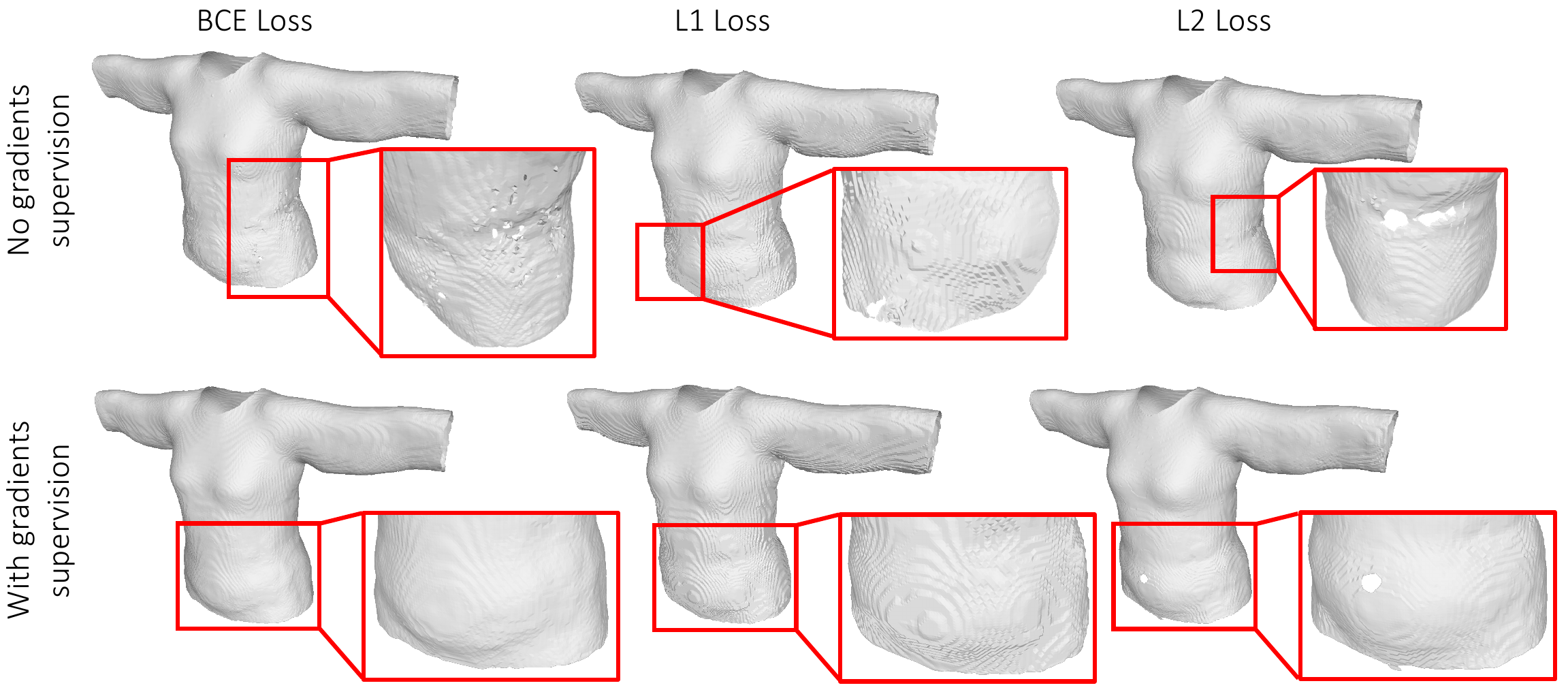}
    \caption{\textbf{Comparison between different loss functions for the garment generative network.} We present the same garment reconstructed by our generative network after being trained for 48 hours with six different alternatives of loss functions.}
    \label{fig:gen_loss_ablation}
\end{figure*}

We report here an ablation study that we conducted to determine the best formulation for $\mathcal{L}_{garm}$, the loss function presented in \cref{eq:gen_loss} of the main paper, that we use to train our garment generative network.

In particular, we consider three variants for $\mathcal{L}_{dist}$, the term of the supervision signal that guides the network to predict accurate values for the garments UDF. In addition to the binary cross-entropy loss (BCE) presented in \cref{eq:gen_dist_loss} of the main paper, we study the possibility of using more traditional regression losses, such as L1 and L2 losses. Adopting the notation introduced in \cref{sec:generative} of the main paper, the L1 loss is defined as $\frac{1}{BN} \sum_{i,j} |min(y_{ij}, \delta) - \widetilde{y}_{ij}|$, while the L2 loss is computed as $\frac{1}{BN} \sum_{i,j} (min(y_{ij}, \delta) - \widetilde{y}_{ij})^2$.

On top of the three variants for $\mathcal{L}_{dist}$, we also consider for each one the possibility of removing the gradients supervision from $\mathcal{L}_{garm}$, \ie, setting $\lambda_g = 0$.

We trained our generative network for 48 hours with the resulting six loss function variants and then compared the quality of the garments reconstructed with the garment decoder. \cref{fig:gen_loss_ablation} presents a significant example of what we observed on the test set. Without gradients supervision (top row of the figure), none of the considered loss functions (BCE, L1 or L2) can guide the network to predict smooth surfaces without artifacts or holes. Adding the gradients supervision (bottom row) induces a strong regularization on the predicted distance fields, helping the network to predict surfaces without holes in most of the cases. However, using the L1 loss leads to rough surfaces, as one can observe in the center column of the bottom row of the figure. The BCE and the L2 losses (first and third columns of the bottom row), instead, produce smooth surfaces that are pleasant to see. We finally opted for the BCE loss over the L2 loss, since the network trained with the latter occasionally predicts surfaces with small holes, as in the example shown in the figure.

\subsection{\texorpdfstring{$\mathcal{L}_{pin}$}{Lpin} for Bottom Garments}

\begin{figure}[ht!]
    \centering
    \includegraphics[width=0.48\textwidth]{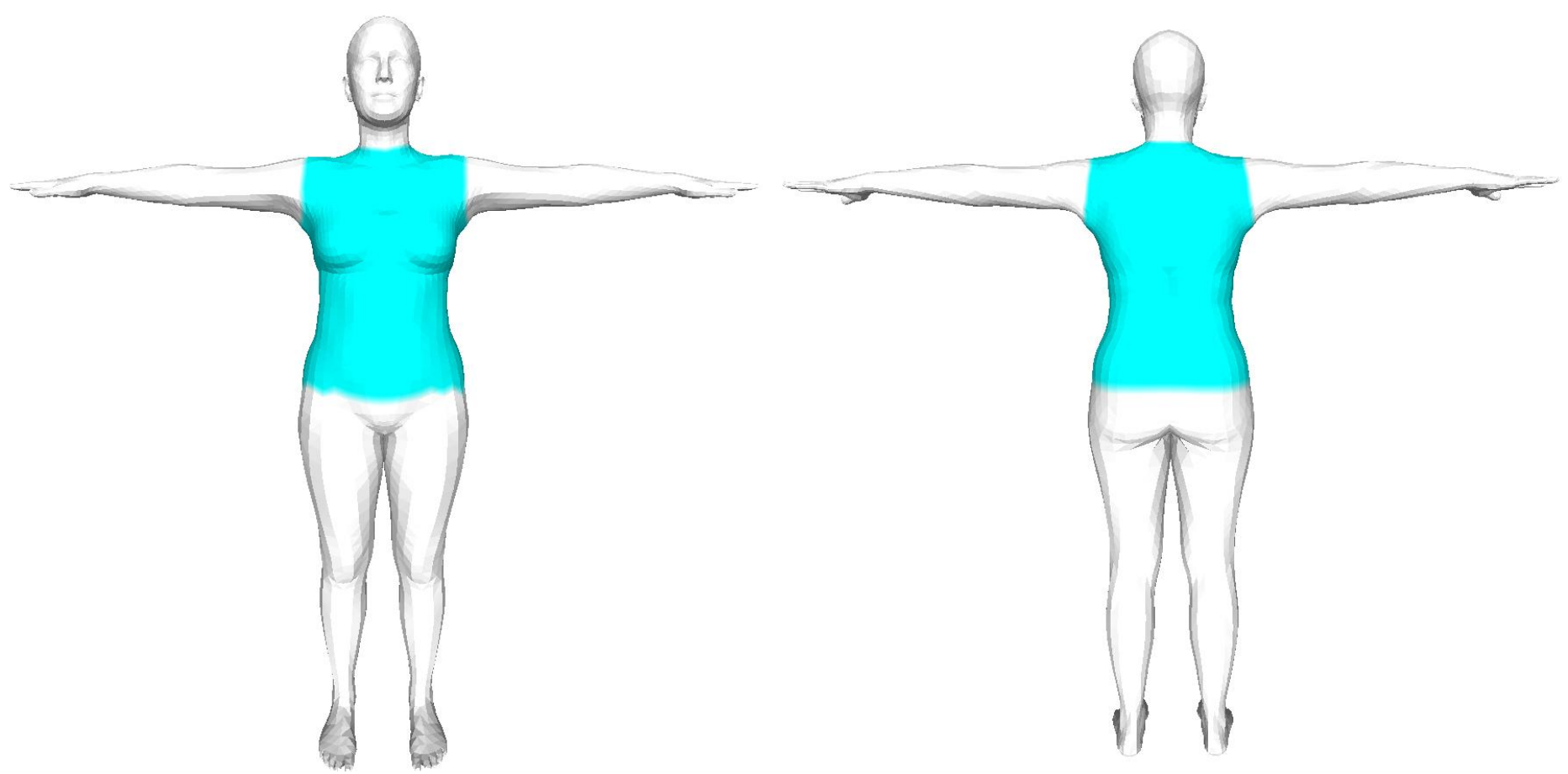}
    \caption{Body region (marked in cyan) used to compute $\mathcal{L}_{pin}$.}
    \label{fig:trunk}
\end{figure}

To determine $V$, the set of bottom garment vertices that need to be constrained by $\mathcal{L}_{pin}$, we first find the closest body vertex $v_B$ for each bottom garment vertex $v$. If $v_B$ locates in the body trunk (cyan region as shown in \cref{fig:trunk}), $v$ is added to $V$.


\begin{figure}[ht!]
    \centering
    \includegraphics[width=0.48\textwidth]{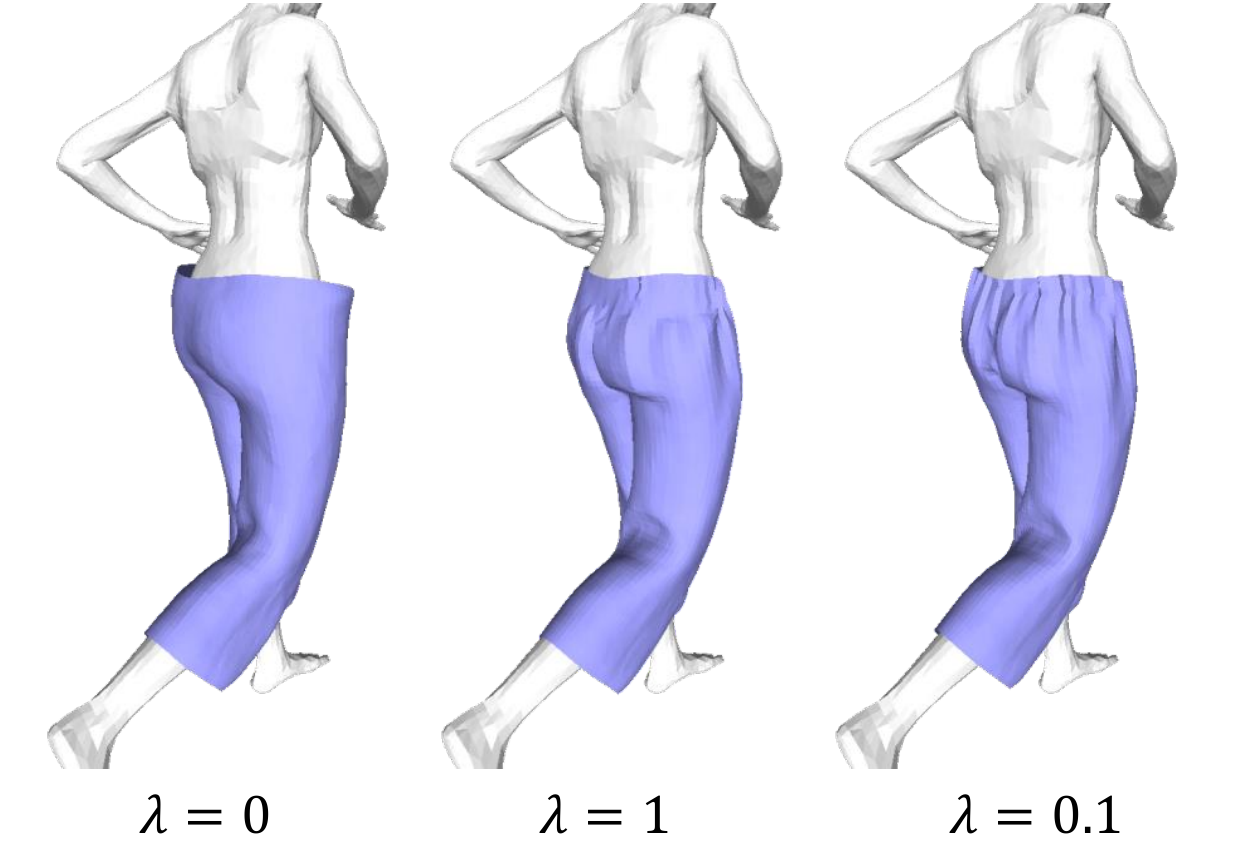}
    \caption{\textbf{Comparison between different values for $\lambda$ of $\mathcal{L}_{pin}$}. To restrict the deformation mainly along the vertical direction (Y axis) and produce natural deformations along other directions, $\lambda$ has to be a positive value smaller than 1. We use $\lambda=0.1$ for our training.}
    \label{fig:supp_pin}
\end{figure}

In \cref{fig:supp_pin}, we show the draping results of bottom garments by using different values for $\lambda$ in $\mathcal{L}_{pin}$. When $\lambda$ equals 0 or 1, the deformations along the X and Z axes are not natural because no constraints or too strong constraints are applied, while it is not the case when $\lambda=0.1$, which is our setting.

\subsection{\texorpdfstring{$\mathcal{L}_{layer}$}{Llayer} for Top-bottom Intersection}
To determine $C$, the set of body vertices covered by both the top and bottom garments, we first subdivide the SMPL body mesh for a higher resolution, and then we compute $C_{top}$ the set of closest body vertices for the given top garment, and $C_{bottom}$ the set of closest body vertices for the bottom. $C$ is derived as the intersection of $C_{top}$ and $C_{bottom}$.


\begin{figure}[ht!]
    \centering
    \includegraphics[width=0.48\textwidth]{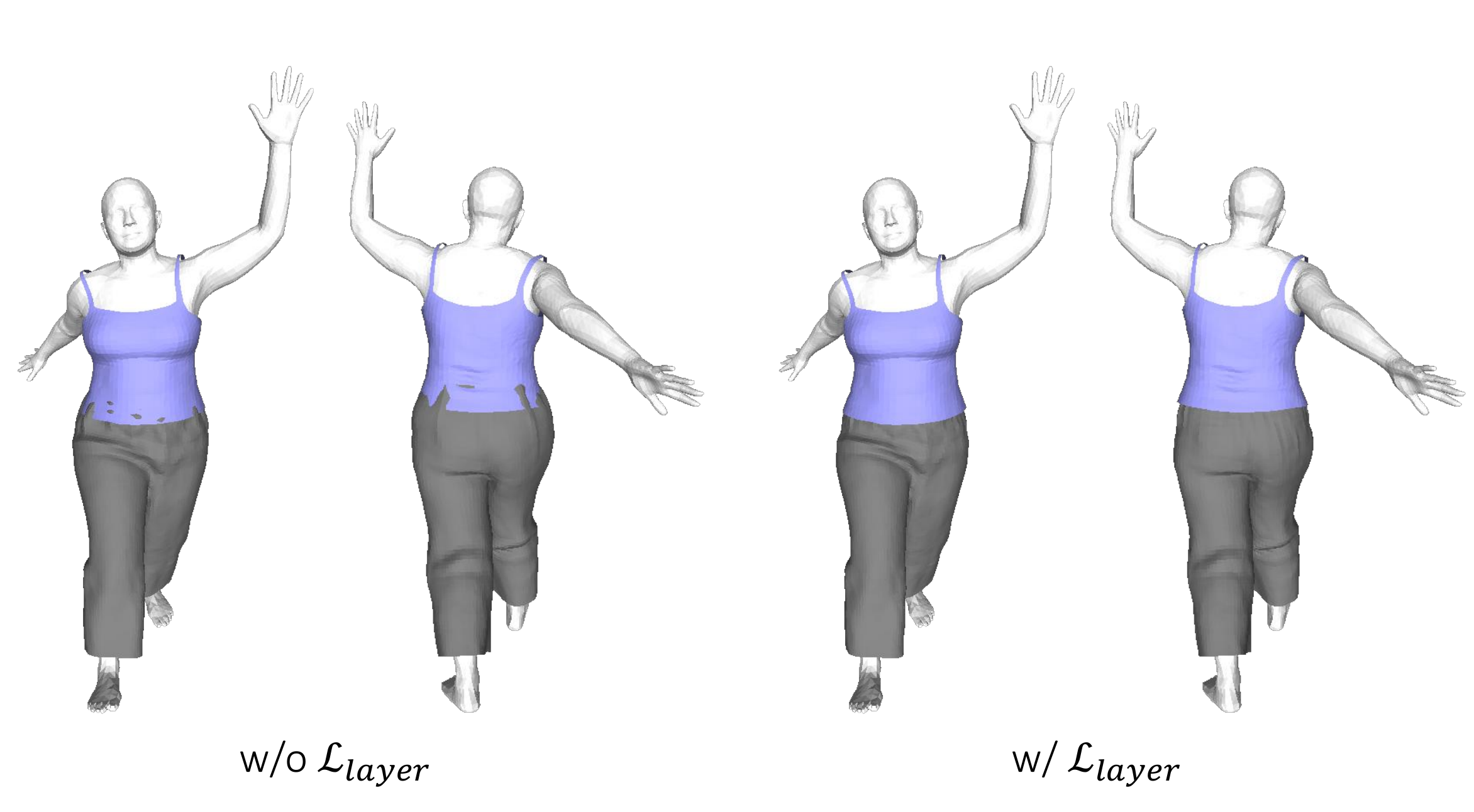}
    \caption{\textbf{Comparison: draping without and with $\mathcal{L}_{layer}$}. Without it, the top and bottom garments intersect with each other.}
    \label{fig:supp_intersection}
\end{figure}
In \cref{fig:supp_intersection} we compare the results of models trained without and with $\mathcal{L}_{layer}$. We can observe that without $\mathcal{L}_{layer}$, the top tank can intersect with the bottom trousers, while it is not the case when using $\mathcal{L}_{layer}$. This indicates the efficacy of $\mathcal{L}_{layer}$ to avoid intersections between garments.

\subsection{Physics-based Refinement}
\label{subsec:supp_refinement}
\begin{figure}[ht!]
    \centering
    \includegraphics[width=0.48\textwidth]{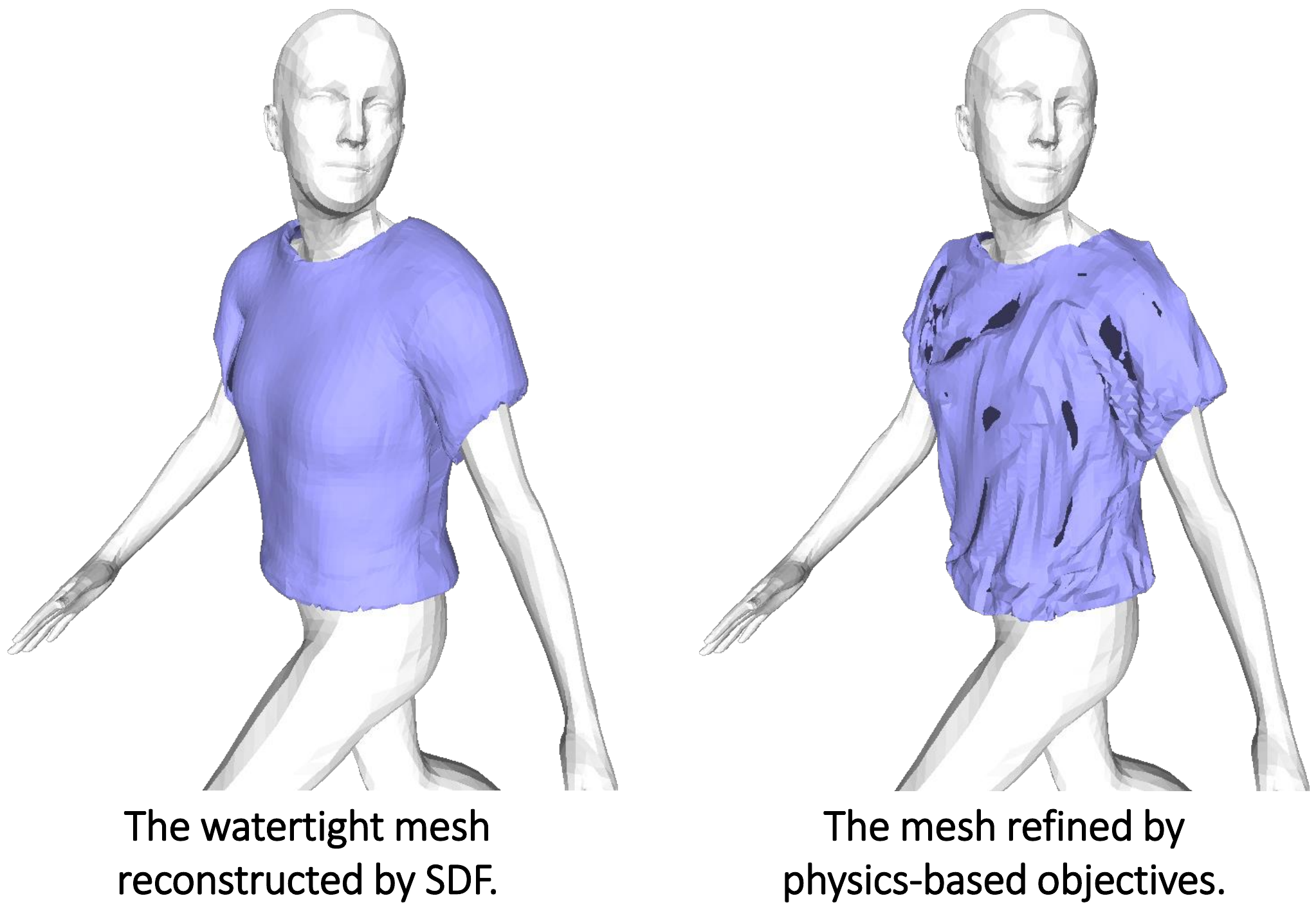}
    \caption{\textbf{Applying post-refinement procedure to watertight mesh}. Left: the watertight mesh reconstructed by DIG~\cite{Li22c}. Right: the same mesh after being refined with physics-based objectives (\cref{eq:post_img}). Physics-based refinement is not compatible with inflated garment meshes, and leads to many self-intersections.}
    \label{fig:supp_sdf}
\end{figure}

After recovering the draped garment $\textbf{G}_D$ 
from images by the optimization of \cref{eq:fit_image} of the main paper, we can apply the physics-based objectives of \cref{eq:physics} (main paper) to increase its level of realism
\begin{equation}
   \scalebox{0.85}{$
   \begin{aligned}
      L(\Delta_\textbf{G}) = & \mathcal{L}_{strain}(\textbf{G}_D+\Delta_\textbf{G}) + \mathcal{L}_{bend}(\textbf{G}_D+\Delta_\textbf{G})\; \label{eq:post_img} \\
      & + \mathcal{L}_{gravity}(\textbf{G}_D+\Delta_\textbf{G}) + \mathcal{L}_{col}(\textbf{G}_D+\Delta_\textbf{G}) \; , 
   \end{aligned}
   $}
\end{equation}
where $\Delta_\textbf{G}$ is the per-vertex-displacement initialized to zero. For the recovery from 3D scans, we apply the following optimization which minimizes both the above physics-based objectives and the Chamfer Distance $d(\cdot)$ to the input scan $\textbf{S}_\textbf{G}$
\begin{equation}
   \scalebox{0.85}{$
   \begin{aligned}
      L(\Delta_\textbf{G}) = & \mathcal{L}_{strain}(\textbf{G}_D+\Delta_\textbf{G}) + \mathcal{L}_{bend}(\textbf{G}_D+\Delta_\textbf{G})\; \label{eq:post_scan} \\
      & + \mathcal{L}_{gravity}(\textbf{G}_D+\Delta_\textbf{G}) + \mathcal{L}_{col}(\textbf{G}_D+\Delta_\textbf{G}) \\
      & + d(\textbf{G}_D+\Delta_\textbf{G}, ~\textbf{S}_\textbf{G})\; . 
   \end{aligned}
   $}
\end{equation}
This refinement procedure is only applicable to open surface meshes, and our UDF model is thus key to enabling it. Applying \cref{eq:post_img} or  \cref{eq:post_scan} to an inflated garment (as recovered by SMPLicit~\cite{Corona21}, ClothWild~\cite{Moon22} and DIG~\cite{Li22c}) indeed yields poor results with many self-intersections as illustrated in \cref{fig:supp_sdf}. This is because inflated garments modelled as SDFs have a non-zero thickness, with distinct inner and outer surfaces whose interactions are not taken into account in this fabric model.
The physics model we apply on garment meshes indeed considers collisions of the garment with the body, but not with itself, which is what happens with the inner and outer surfaces in \cref{fig:supp_sdf}. Adding a physics term to prevent self intersections would not be trivial, and is related to the complex task of untangling layered garments~\cite{Santesteban22b,Buffet19}

Note that this is also the case for most garment draping softwares~\cite{Narain12,Narain13,Pfaff14,Tang18d,Gundogdu22} to expect single layer garments. Modeling garment with UDFs is thus a key feature of our pipeline for its integration in downstream tasks.

Both the optimizations of \cref{eq:fit_image,eq:fit_scan} of the main paper and \cref{eq:post_img,eq:post_scan} are done with Adam \cite{Fan17a} but with different learning rates set to 0.01 and 0.001 respectively.

\section{Additional Results}
\label{sec:supp_results}
\subsection{Garment Encoder/Decoder}
%

\begin{figure*}
    \centering
    \includegraphics[width=0.99\textwidth,trim={1mm 1mm 0 1mm},clip]{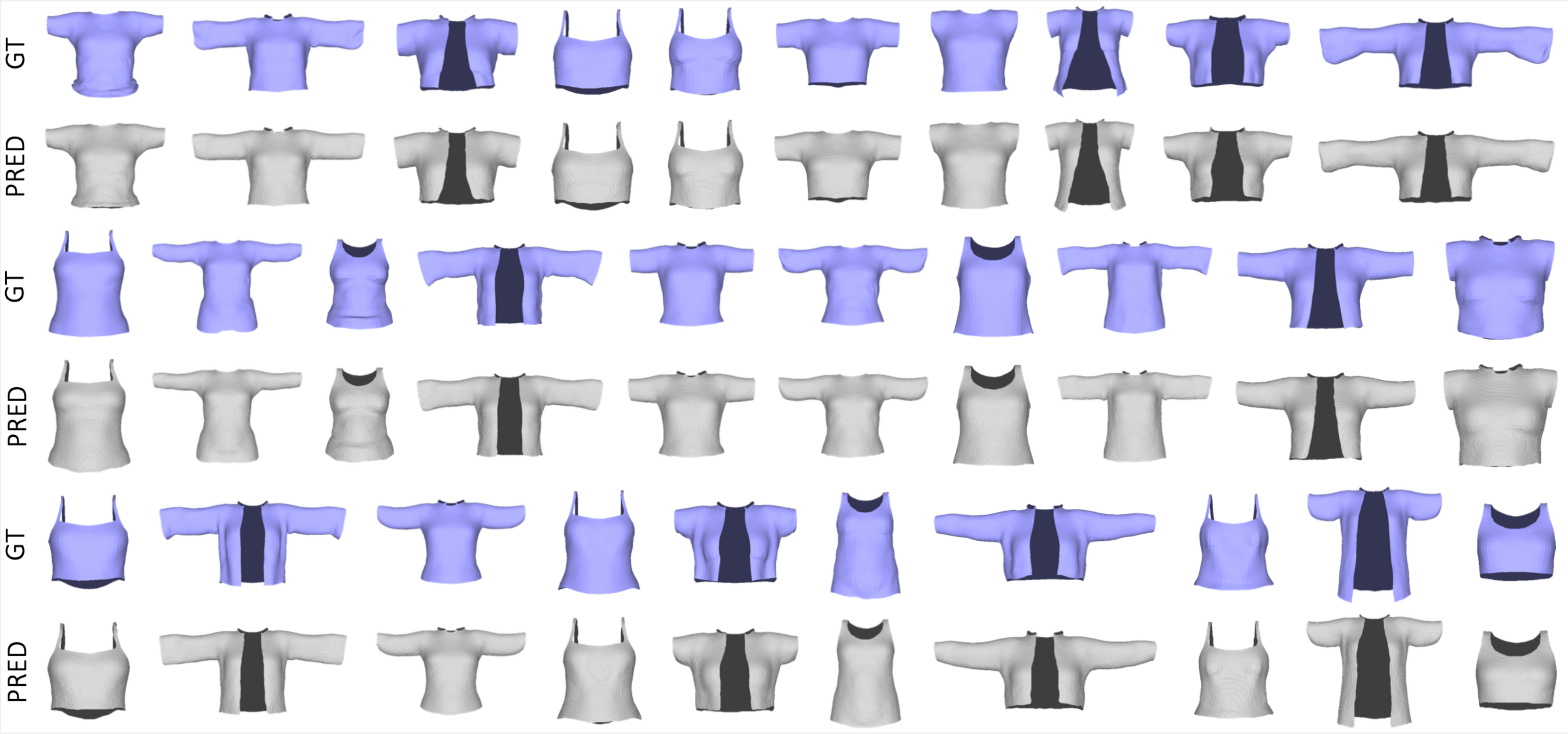}
    \caption{\textbf{Generative network: reconstruction of unseen garments in neutral pose/shape (top garments).} Latent codes for unseen garments can be obtained with our garment encoder. These codes are then used by the garment decoder to reconstruct open surface meshes. Input garments are colored in purple, while the reconstructed meshes are colored in gray.}
    \label{fig:top_rec_all}
\end{figure*}


\begin{figure*}
    \centering
    \includegraphics[width=0.99\textwidth,trim={1mm 0 0 1mm},clip]{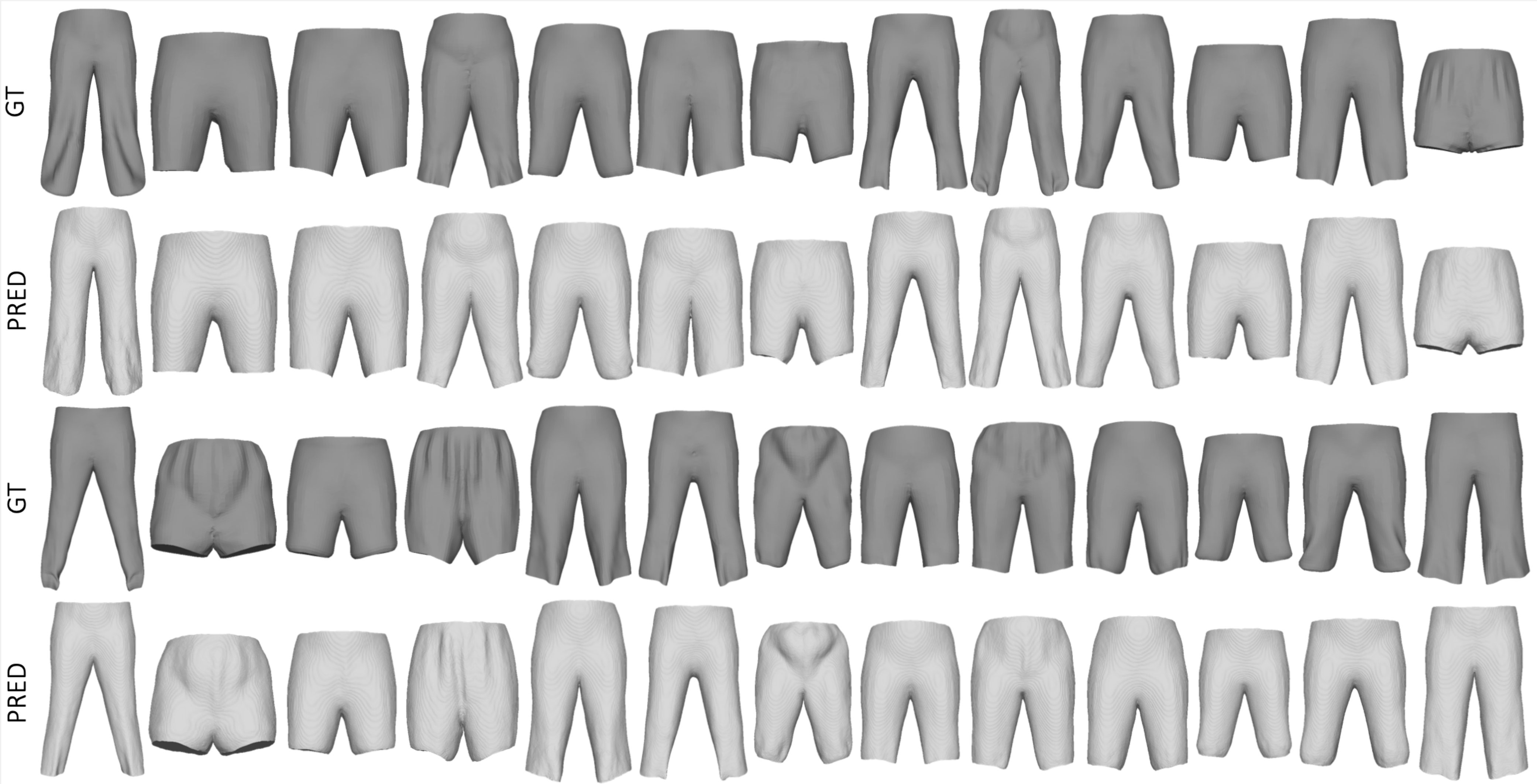}
    \caption{\textbf{Generative network: reconstruction of unseen garments in neutral pose/shape (bottom garments).} Latent codes for unseen garments can be obtained with our garment encoder. These codes are then used by the garment decoder to reconstruct open surface meshes. Input garments are colored in dark gray, while the reconstructed meshes are colored in light gray.}
    \label{fig:bottom_rec_all}
\end{figure*}


\begin{figure*}
    \centering
    \includegraphics[width=0.95\textwidth,trim={0 1mm 1mm 1mm},clip]{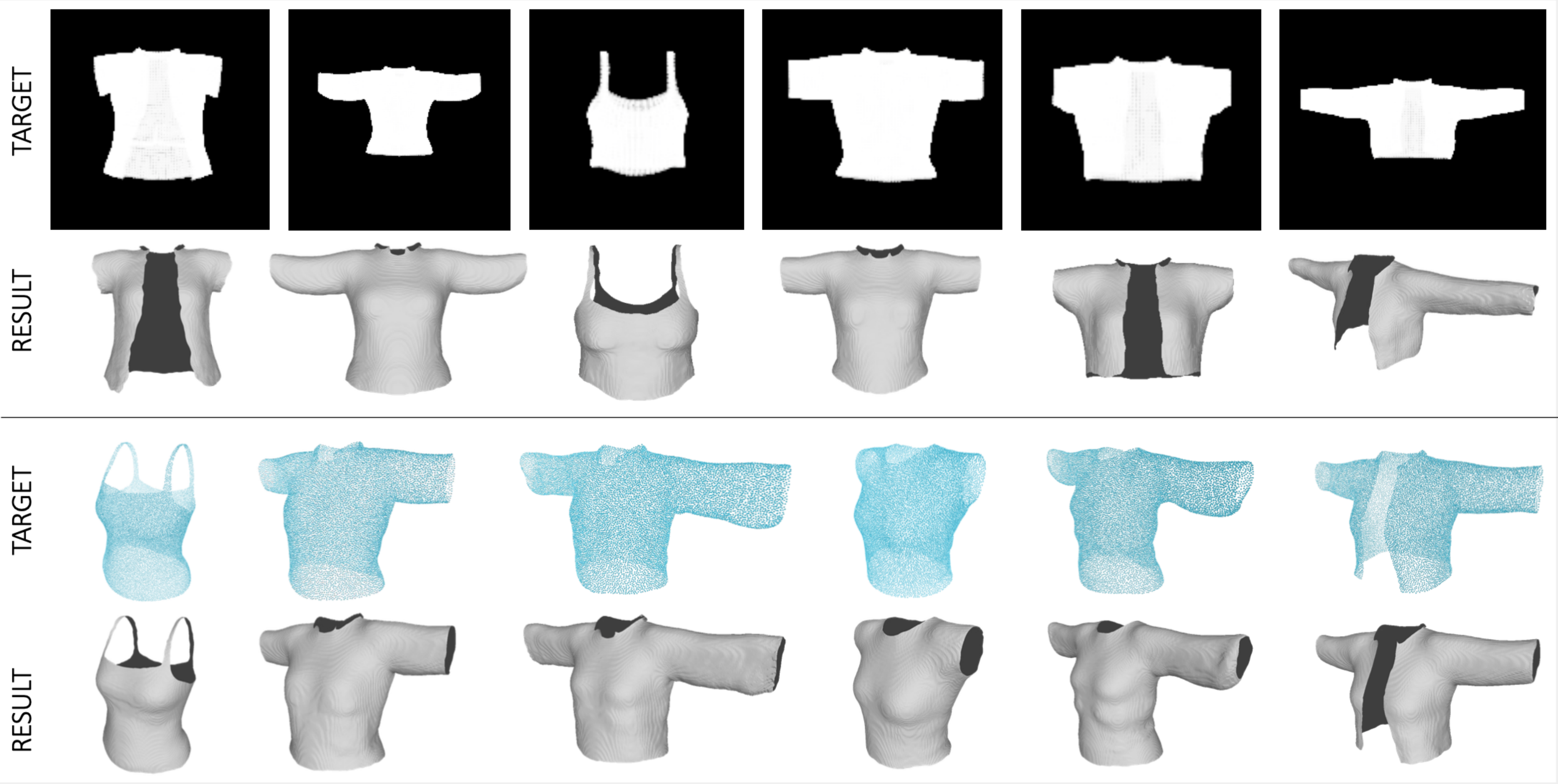}
    \caption{\textbf{Generative network: latent space optimization (top garments).} After training, we can explore the latent space learned by the garment generative network with gradient descent, to recover target garments from 2D silhouettes (top) or 3D point clouds (bottom).}
    \label{fig:gen_lso_top}
\end{figure*}


\begin{figure*}
    \centering
    \includegraphics[width=0.95\textwidth,trim={1mm 1mm 0 0},clip]{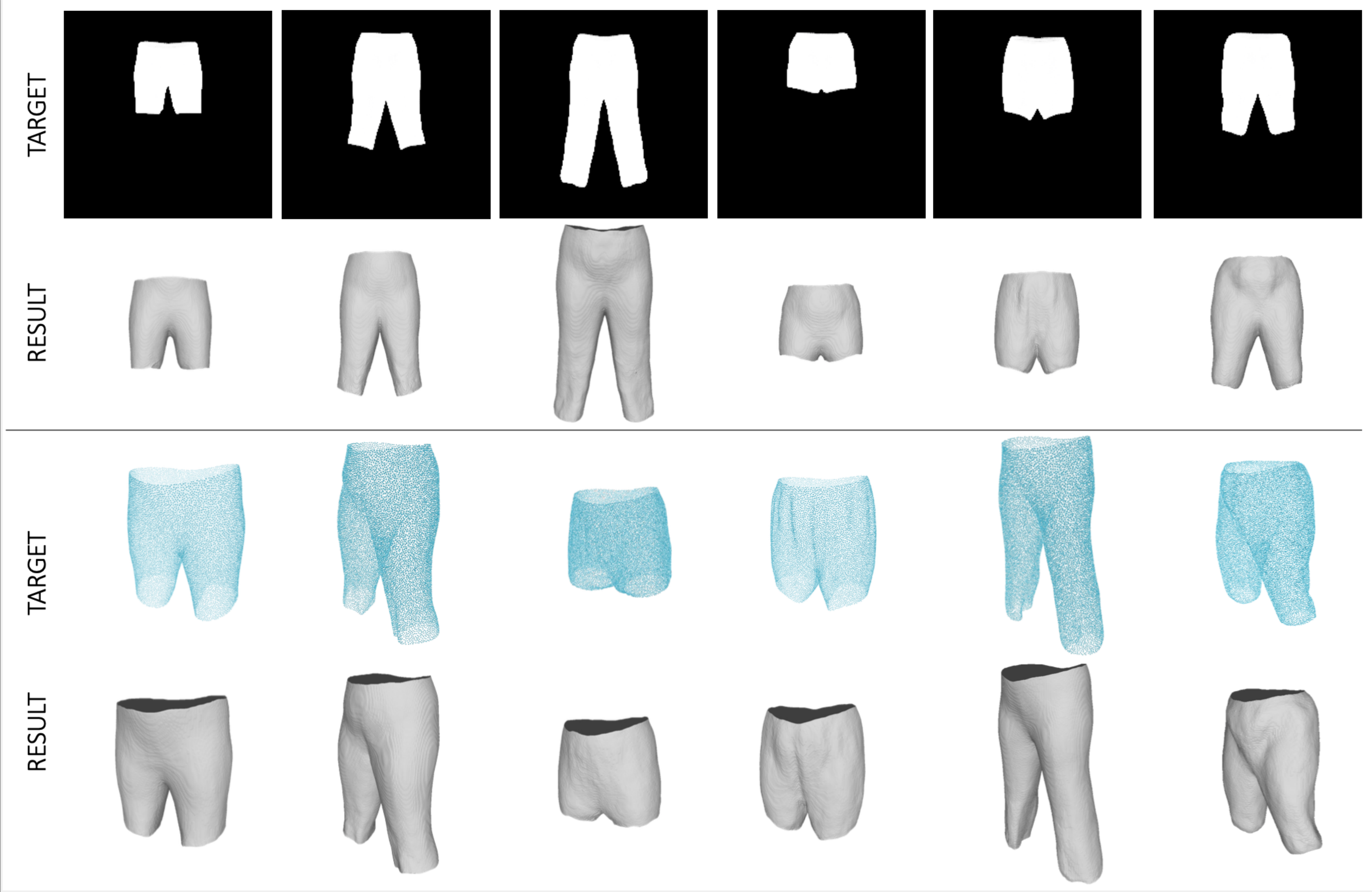}
    \caption{\textbf{Generative network: latent space optimization (bottom garments).} After training, we can explore the latent space learned by the garment generative network with gradient descent, to recover garments from 2D silhouettes (top) or 3D point clouds (bottom).}
    \label{fig:gen_lso_bottom}
\end{figure*}

\subsubsection{Additional Qualitative Results}
\cref{fig:top_rec_all} and \cref{fig:bottom_rec_all} show the encoding-decoding capabilities of our garment generative network for top and bottom \textit{test} garments, respectively. The ground-truth garments are passed through the garment encoder, which produces a compact latent code for each clothing item. Then, our garment decoder reconstructs the input garments surface from the latent codes.
It is possible to notice how the output garments closely match the input ones, both in terms of geometry and topology.


\begin{figure*}[ht!]
    \centering
    \includegraphics[width=0.99\textwidth]{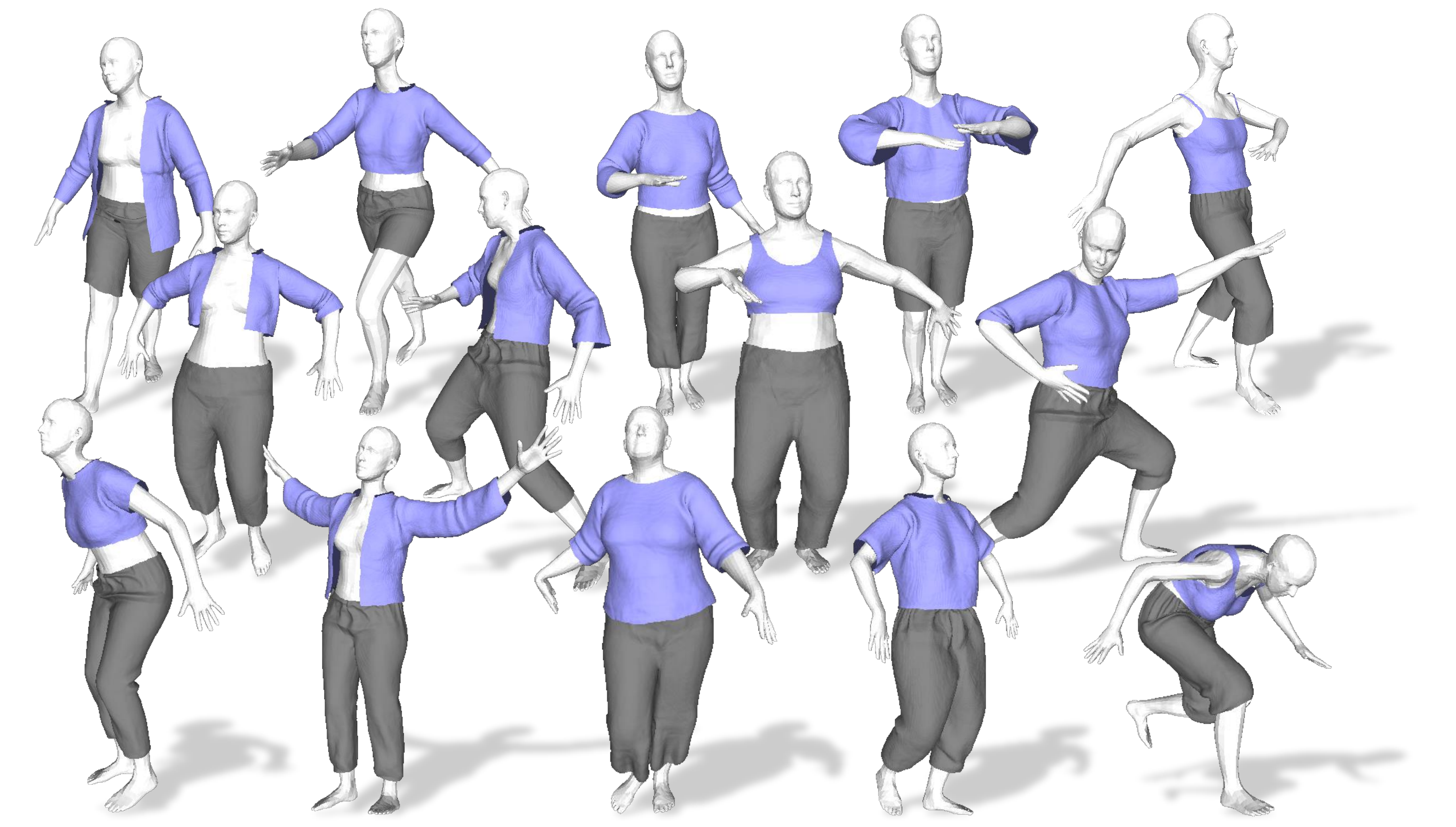}
    \caption{\textbf{Additional results:} draping garments of different topologies over bodies in various shapes and poses with our method.}
    \label{fig:supp_drape}
\end{figure*}

\subsubsection{Latent Space Optimization (LSO).}
After training the garment generative network, we obtain a latent space that allows us to sample a garment latent code and to feed it to the implicit decoder to reconstruct the explicit surface. We study here the possibility of exploring the garment latent space by the means of LSO. To do that, given a target 2D silhouette or a sparse 3D point cloud of a garment, we optimize with gradient descent a latent code -- initialized to the training codes average -- so that the frozen decoder conditioned on it can produce a garment which fits the target image or point cloud.

Given the silhouette $\mathcal{S}$ of a target garment, we can retrieve its latent code $\mathbf{z}$ by minimizing
\begin{equation}
\scalebox{0.9}{$
\begin{aligned}
    L(\bz) &= L_{\text{IoU}}(R(\mathbf{G}), \mathcal{S})\; , \label{eq:gen_lso_image} \\
    \mathbf{G} &= \text{MeshUDF}(D_G(\cdot,\bz)) \; , 
\end{aligned}
$}
\end{equation}
where $L_{\text{IoU}}$ is the IoU loss \cite{Li21g} in pixel space measuring the difference between 2D silhouettes , $R(\cdot)$ is a differentiable silhouette renderer for meshes~\cite{Pytorch3D}, and $\textbf{G}$ is the garment mesh reconstructed with our garment decoder using $\bz$.

In the case of a target garment provided as a point cloud $\mathcal{P}$, the garment latent code $\bz$ can be obtained by minimizing
\begin{equation}
\scalebox{0.9}{$
\begin{aligned}
    L(\bz) &= d(ps(\textbf{G}), \mathcal{P})\; , \label{eq:gen_lso_pcd} \\
    \textbf{G} &= \text{MeshUDF}(D_G(\cdot,\bz)) \; , 
\end{aligned}
$}
\end{equation}
where $d(a,b)$ is the Chamfer distance \cite{Fan17a} between point clouds $a$ and $b$, and $ps(\cdot)$ represents a differentiable procedure to sample points from a given mesh \cite{Pytorch3D}.

In both cases, we run the optimization for 1000 steps, with Adam optimizer \cite{Kingma15} and learning rate set to $0.01$.

In \cref{fig:gen_lso_top} and \cref{fig:gen_lso_bottom} we present some results of the LSO procedures here described, showing that the latent space learned by the garment generative network can be explored effectively with gradient descent to recover the codes associated with the target garments.

\subsection{Draping Network}
\subsubsection{Additional Qualitative Results}
In \cref{fig:supp_drape} we show additional qualitative results of garment draping produced by our method, where the garment meshes are generated by our UDF model. It can be seen that our method can realistically drape garments with different topologies over bodies of various shapes and poses.

\subsubsection{Euclidean Distance is not a Good Metric}
\begin{figure}
    \centering
    \includegraphics[width=0.48\textwidth]{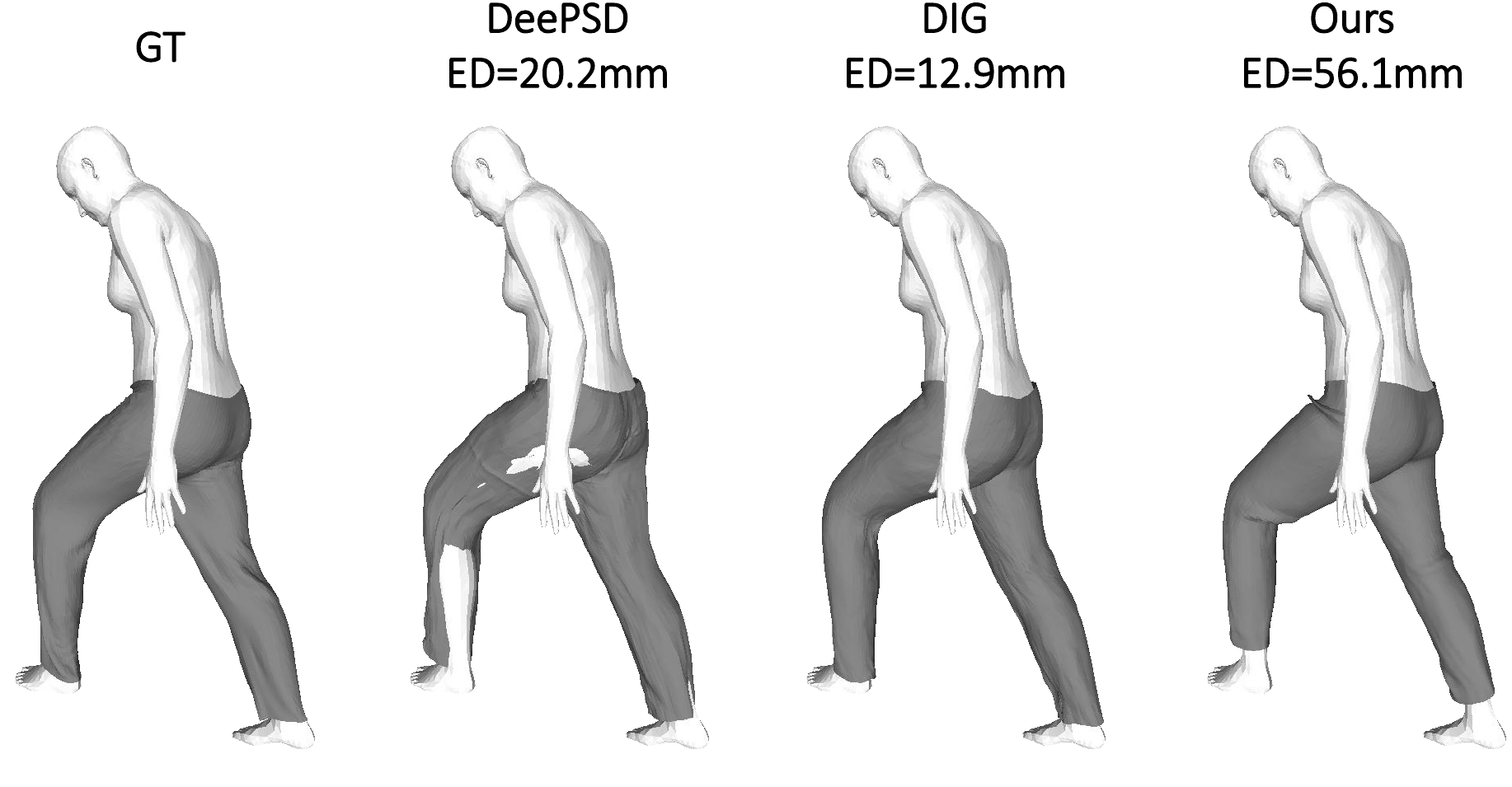}
    \caption{\textbf{Comparison between DeePSD, DIG and our method.} Our result is more realistic than the others despite having the highest Euclidean distance (ED) error.}
    \label{fig:supp_ED}
\end{figure}

In \cref{fig:supp_ED}, we show an example of bottom garment where our result is more realistic than the competitors DeePSD~\cite{Bertiche21} and DIG~\cite{Li22c} despite having the highest Euclidean distance. This demonstrates again that Euclidean distance is not able to measure the draping quality, as discussed in the main paper.

\subsubsection{Quantitative Evaluation in Physics-based Energy}

\begin{table*}[ht!]
    \begin{center}
      \scalebox{.9}{
        \begin{tabular}{c | c | c | c | c}
        \toprule
        Top & Strain $\downarrow$ & Bending $\downarrow$ & Gravity $\downarrow$ & Total $\downarrow$ \\
        \midrule
         DeePSD & 7.22  & \textbf{0.01}  & \textbf{0.98}  & 8.21 \\
         DIG    & 6.32  & \textbf{0.01}  & 1.05  & 7.38 \\
         \midrule
         Ours  & \textbf{0.43} & \textbf{0.01} & 1.05 & \textbf{1.81} \\
        \bottomrule
        \end{tabular}
      }
      \scalebox{.9}{
        \begin{tabular}{c | c | c | c | c}
        \toprule
        Bottom & Strain $\downarrow$ & Bending $\downarrow$ & Gravity $\downarrow$ & Total $\downarrow$ \\
        \midrule
         DeePSD   & 8.46  & 0.02  & 0.90  & 9.38 \\
         DIG      & 7.48  & \textbf{0.01}  & 0.90  & 8.39 \\
         \midrule
         Ours     & \textbf{0.41} & \textbf{0.01} & \textbf{0.86} & \textbf{1.28} \\
        \bottomrule
        \end{tabular}
        }
      \end{center}
      \caption{\textbf{Draping unseen garment meshes.} Quantitative comparison in physics-based energy between DeePSD, DIG and our method. ``Strain'', ``Bending'' and ``Gravity`` denote the membrane strain energy, the bending energy and the gravitational potential energy, respectively.}
      \label{tab:energy}
\end{table*}
In \cref{tab:energy}, we report the physics-based energy of \textit{Strain}, \textit{Bending} and \textit{Gravity} as proposed by~\cite{Santesteban22} on test garment meshes when draped by DeePSD, DIG and our method. These energy terms are used as training losses for our garment network (\cref{eq:physics,,eq:pin} of the main paper).
For the gravitational potential energy, we choose the lowest body vertex as the 0 level. Generally, our results have the lowest energies, especially for the \textit{Strain} component. Since DeePSD and DIG do not apply constraints on mesh faces, their results exhibit much higher \textit{Strain} energy. This indicates that our method can produce results that have more realistic physical properties. 

\subsection{Inference Times}
We report inference times for the components of our framework, computed on an NVIDIA Tesla V100 GPU.
The garment encoder, which needs to be run only once for each garment, takes $\sim$25 milliseconds. The decoder takes $\sim$2 seconds to reconstruct an explicit garment mesh from a given latent code, including the modified Marching Cubes from~\cite{Guillard22b} at resolution $256^3$.

The draping network takes $\sim$5 ms to deform a garment mesh composed of 5K vertices. Since it is formulated in an implicit manner and is queried at each vertex, its inference time increases to $\sim$8 ms for a mesh with 8K vertices, or $\sim$53 ms with 100K vertices.

\subsection{Fitting SMPLicit~\cite{Corona21} to 3D Scans}

\begin{figure}[ht!]
    \centering
    \includegraphics[width=0.48\textwidth]{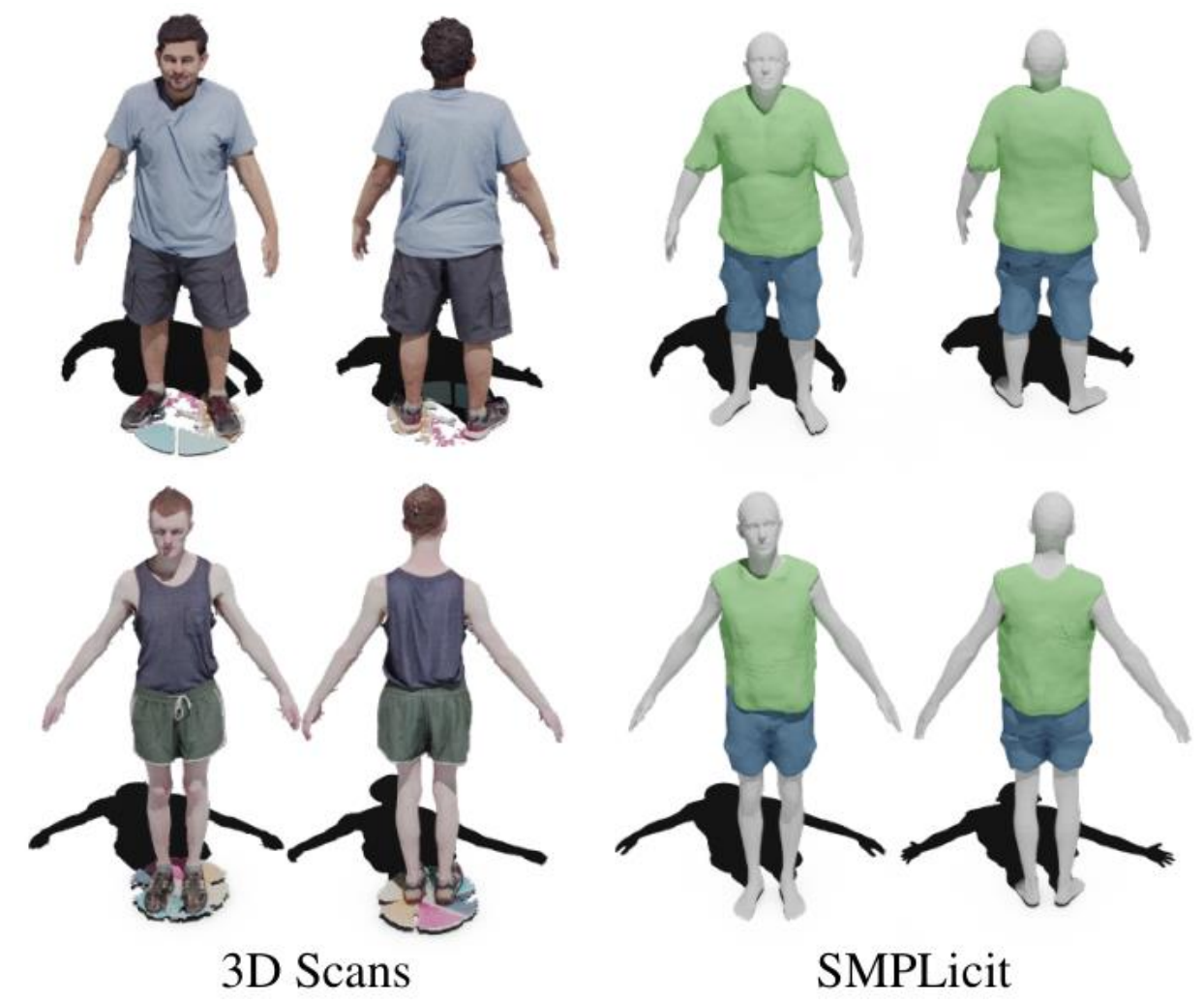}
    \caption{\textbf{Recovered garments of SMPLicit from 3D scans}. Figures are extracted from \cite{Corona21}.}
    \label{fig:supp_scan}
\end{figure}

In \cref{fig:supp_scan} we show results of fitting the concurrent approach SMPLicit~\cite{Corona21} to 3D scans of the SIZER dataset~\cite{Tiwari20}. We can observe that they are not as realistic as ours shown in \cref{fig:scan} of the main paper. Since we have no access to their code and not enough information for a re-implementation, we directly extract this figure from~\cite{Corona21}.

\section{Human Evaluation}
\label{sec:supp_human_eval}

\begin{figure*}
    \centering
    \fbox{\includegraphics[width=0.96\textwidth]{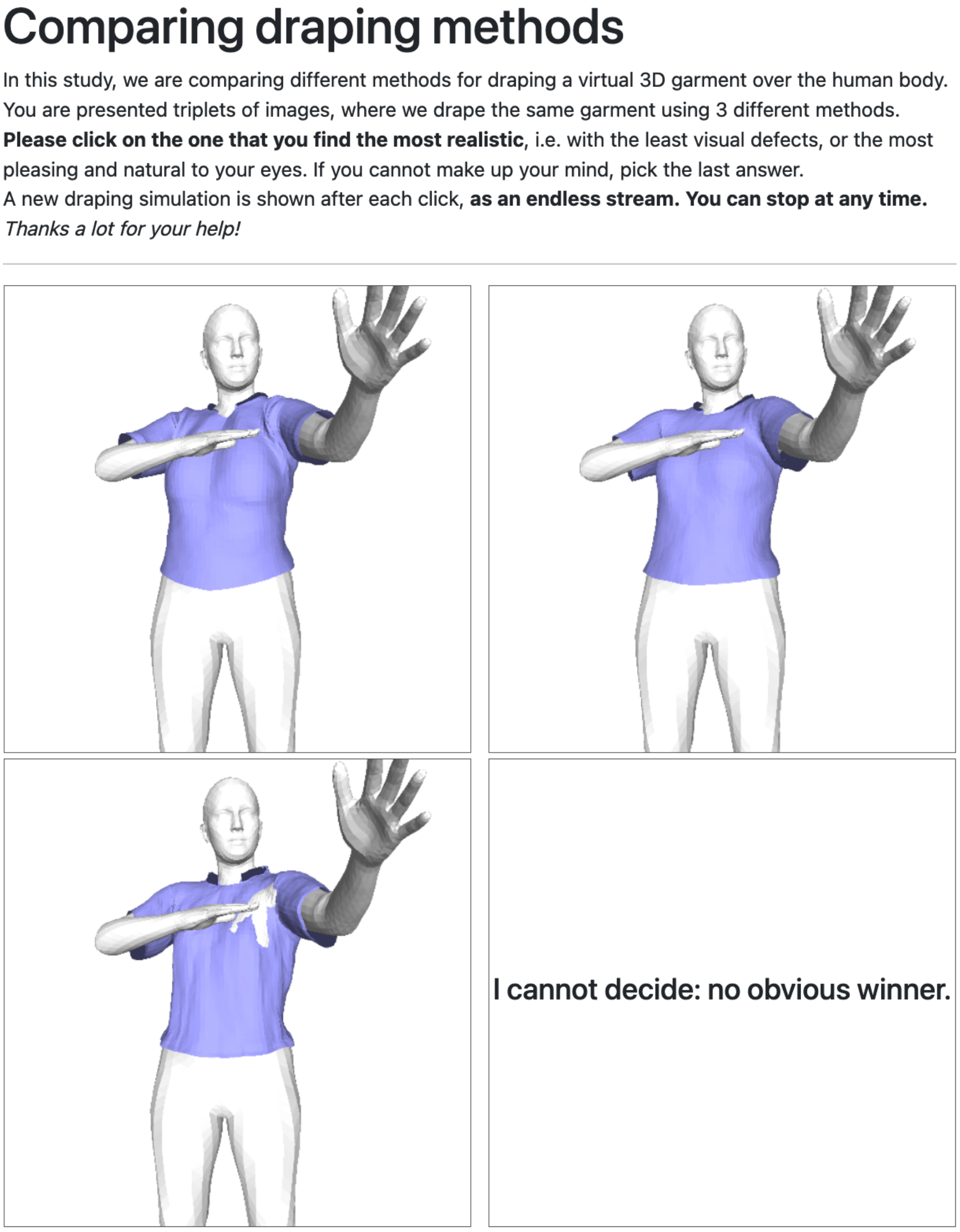}}
    \caption{\textbf{Interface of our qualitative survey.} The garment is draped with our method, DIG, and DeePSD, in a random order.}
    \label{fig:supp_human_eval}
\end{figure*}

In \cref{fig:supp_human_eval} we show the interface and instructions that were presented to the 187 respondents of our survey. These evaluators were volunteers with various backgrounds from the authors respective social circles, which were purposely not given any further detail or instruction.
We collected collected 3738 user opinions in total, each user expressing 20 opinions on average.

\end{document}